\PassOptionsToPackage{hidelinks}{hyperref}
\documentclass{iopjournal-arxiv}

\newcommand{\macro}[1]{\texttt{\textbackslash#1}}
\newcommand{\m}[1]{\macro{#1}}

\usepackage{graphicx,amsmath,amssymb,amsthm}
\usepackage{xcolor}
\usepackage{booktabs}
\usepackage{color}
\newcommand{\expval}[1]{\left\langle #1 \right\rangle}
\newcommand{\pdv}[2]{\frac{\partial #1}{\partial #2}}
\usepackage{verbatim}
\usepackage[caption=false]{subfig}
\usepackage{ifthen}
\usepackage{caption}
\usepackage{subcaption} 

\usepackage{ragged2e}
\justifying
\AtBeginDocument{\justifying}

\usepackage{multirow}
\usepackage{makecell}

\hypersetup{
colorlinks,
linkcolor={blue!80!black},
citecolor={blue!80!black},
urlcolor={blue!80!black},
linktoc=page
}

\newcommand\beq{\begin{equation}}
\newcommand\eeq{\end{equation}}

\renewcommand{\b}[1]{\mathbf{#1}}
\renewcommand{\c}[1]{\mathcal{#1}}
\renewcommand{\u}{\uparrow}
\renewcommand{\d}{\downarrow}

\newcommand{\be}{\begin{equation}}
\newcommand{\ee}{\end{equation}}

\newcommand{\N}{\mathcal{N}}

\makeatletter
\g@addto@macro{\UrlBreaks}{\do\a\do\b\do\c\do\d\do\e\do\f\do\g\do\h\do\i\do\j\do\k\do\l\do\m\do\n\do\o\do\p\do\q\do\r\do\s\do\t\do\u\do\v\do\w\do\x\do\y\do\z\do\A\do\B\do\C\do\D\do\E\do\F\do\G\do\H\do\I\do\J\do\K\do\L\do\M\do\N\do\O\do\P\do\Q\do\R\do\S\do\T\do\U\do\V\do\W\do\X\do\Y\do\Z\do\0\do\1\do\2\do\3\do\4\do\5\do\6\do\7\do\8\do\9}
\makeatother
\begin{document}

\fancyhead[R]{Mirafzali {\it et al}\ }

\articletype{Paper}

\title{Holographic generative flows with AdS/CFT}

\author{Ehsan Mirafzali$^{1,\dagger,*}$, Sanjit Shashi$^{1,2,\dagger,*}$, Sanya Murdeshwar$^1$, Edgar Shaghoulian$^2$, Daniele Venturi$^3$ and Razvan Marinescu$^1$}

\affil{$^1$Department of Computer Science and Engineering, University of California, Santa Cruz, California, USA}

\affil{$^2$Santa Cruz Institute for Particle Physics, Department of Physics, University of California, Santa Cruz, California, USA}

\affil{$^3$Department of Applied Mathematics, University of California, Santa Cruz, California, USA}

\affil{$^\dagger$These authors contributed equally to this work.}

\affil{$^*$Authors to whom any correspondence should be addressed.}

\email{smirafza@ucsc.edu, sashashi@ucsc.edu}


\begin{abstract}\justifying
Holography, in the form of the anti-de Sitter/conformal field theory (AdS/CFT) correspondence, offers a natural setting for generative modelling. Data on a boundary manifold lifts into a higher-dimensional bulk through a propagator, and this extra dimension plays the role of a flow parameter. We exploit this structure to build GenAdS, an approach to generative flow matching in which the dynamics are represented by the evolution of fields in AdS, together with a residual correction learned by a neural network. Boundary samples are encoded as scalar sources, transported into the bulk along the flow, and decoded after numerical integration. Our paradigm combines a Fourier-space encoding scheme for the data as AdS sources, a normalised radial phase space in which to stage the flow-matching dynamics, and a Klein--Gordon backbone to guide the flow. On a two-dimensional checkerboard benchmark, our experiments show that most of the benefit of GenAdS comes from the Fourier representation and convolutional architecture. However, when we remove momentum-channel regularisation, our most physics-informed GenAdS variant rivals the strongest physics-free control on boundary violation. On MNIST, GenAdS models remain close to a convolutional baseline on fidelity while achieving significantly higher recall at comparable precision, suggesting a fidelity-coverage trade-off. Our findings establish GenAdS as a physically interpretable and experimentally controllable framework for generative modelling, with many avenues for future extension.
\end{abstract}
\thispagestyle{plain}
\section{Introduction}

Generative machine learning has developed rapidly in recent years. Stochastic models, in particular diffusion \cite{sohldickstein2015deepunsupervisedlearningusing,ho2020denoisingdiffusionprobabilisticmodels}, remain prominent, but deterministic flow-based models have emerged as a serious alternative. The early framework was that of continuous normalising flows \cite{chen2019neuralordinarydifferentialequations,grathwohl2018ffjordfreeformcontinuousdynamics}, in which a learned velocity field defines an ordinary differential equation and a log-likelihood loss closes the training loop. The price is paid in computation, since training a continuous normalising flow requires simulation of the flow itself, which is slow and costly. Flow matching \cite{lipman2023flowmatchinggenerativemodeling} addresses this difficulty by replacing the simulation-based objective with a direct one and has therefore become a natural starting point for new flow-based models.

Such models are entirely data-driven and typically ignore any structure the data may possess. One way to repair this is to bring physics in as inductive bias \cite{Raissi2019,pmlr-v168-djeumou22a}. The approach is natural for flows. A flow is a velocity field, and physical equations of motion supply velocity fields with built-in symmetries, conservation laws, and analytic structure. The recent GenPhys framework \cite{Liu:2023iee} pursues this idea systematically, treating a wide class of partial differential equations as engines for generative flows. A complementary line of thought situates the flow itself in an extra dimension, recasting generation of $d$-dimensional data as geometry in $d{+}1$ dimensions and opening the door to non-Euclidean machine learning (cf. \cite{geomflow:2023}).

These two ideas, taken together, are precisely the situation invoked by the holographic principle of quantum gravity \cite{tHooft:1993dmi,Susskind:1994vu}, in which a $(d{+}1)$-dimensional theory of gravity is dual to a non-gravitational field theory on a $d$-dimensional boundary. The bulk geometry emerges from boundary data, much as a hologram does from its source. The most precise realisation of this principle is the anti-de Sitter/conformal field theory (AdS/CFT) correspondence \cite{Maldacena:1997re,Witten:1998qj}. AdS is a negatively curved space, the higher-dimensional analogue of the Poincar\'e disk; a CFT is a field theory invariant under scale transformations. In AdS/CFT, the CFT lives on the boundary and the bulk physics defines a flow of $d$-dimensional theories along the radial direction \cite{Balasubramanian:1999jd,deBoer:1999tgo}.

The translation to machine learning is direct, though it must be phrased with care. Training samples are identified with boundary-source profiles for a scalar operator of the boundary theory. The data distribution then induces a measure on the space of sources, and the generative model samples from this measure, thereby generating sources (as opposed to CFT field configurations). The radial direction, identified in AdS/CFT with the renormalisation-group (RG) scale of the boundary theory \cite{deBoer:1999tgo,Verlinde:1999xm}, plays the role of a flow parameter. A related analogy connects AdS/CFT to quantum error correction \cite{Swingle:2009bg,VanRaamsdonk:2010pw,Pastawski:2015qua}, and a more direct one casts AdS physics as a deep-learning network \cite{Hashimoto:2018ftp}, using machine learning to recover holographic structure. Our programme runs in the opposite direction. Rather than using machine learning to infer holography, we use holography as an inductive bias for ordinary machine-learning tasks. The two directions are complementary rather than mutually exclusive; one uses data to learn a holographic map, while the other fixes an analytic holographic map and asks what it contributes to learning.

Concretely, we build a flow-matching generative model whose dynamics are governed by AdS/CFT physics (cf. \cite{Zhdanov:2025xkm}). We pick a particular bulk theory, namely Klein--Gordon scalar theory in AdS, and treat boundary CFT sources as proxies for ground-truth data. The flow then follows Klein--Gordon dynamics (represented in Fourier space) supplemented by a residual correction learned by a convolutional neural network.\footnote{We do this because Klein--Gordon scalars obey linear dynamics, so the resulting encoding carries only two-point (Gaussian) information about the source. Non-Gaussian features must be supplied by a learned residual.} This combination preserves the analytic structure of AdS while retaining enough flexibility to generate datasets that have nothing to do with physics. To bring raw data into this field-theoretic setting, we develop a notion of holographic encoding. Training samples are treated as boundary sources and projected into the bulk by the AdS propagator. We call the resulting framework \textbf{Gen}erative \textbf{AdS} (GenAdS).\footnote{The code is available at our \href{https://github.com/LVCIFERI/Holographic-Generative-Flows-with-AdS-CFT---Planar}{GitHub repository}.} None of the ingredients are new; analytic bulk-to-boundary propagators, flow matching, and physics-plus-residual decompositions all have established literatures, as reviewed below. What is novel is the combination of all of these elements.

The experiments are all organised around measuring the relative contributions of each of these ingredients to the efficacy of GenAdS; we briefly summarise them now. In the first set of experiments, we train models to generate the two-dimensional checkerboard. In addition to several GenAdS variants, each of which incorporates different amounts of the physics, we also train two parameter-matched controls, a fully connected neural network and a convolutional neural network, on this task. The GenAdS variants substantially outperform the former, but not the latter, showing that the improvements are attributable to the Fourier representation and convolutional architecture rather than AdS-specific structure. Also, using generic low-pass filters (heat-kernel and Mat\'ern) rather than the AdS propagator in the holographic encoding does not yield significant change in our metrics. However, these experiments all utilise a noise term on the momentum channel for the purpose of regularisation. An ablation of this noise reveals that it suppresses our most physics-informed GenAdS model, and setting noise to zero allows this model to perform comparably to our best physics-free control on boundary violation.

We also test the ability of GenAdS to generate the Modified National Institute of Standards and Technology (MNIST) dataset \cite{lecun2002}. On this dataset, the GenAdS models remain close to a strong convolutional baseline on Fr\'echet inception distance, which measures fidelity, while also achieving much higher recall. Hence, the encoding delivers measurably better coverage of the data manifold at a small cost in Fr\'echet statistics.

Our main claim of GenAdS is not that holography beats physics-free networks. Rather, we are presenting GenAdS as a working implementation of holography-inspired generative modelling, with interpretable inductive bias and significant room for refinement and future development.

\paragraph{Related work} We now review four adjacent lines of research that frame our work in the literature.

\smallskip\noindent\emph{Holography learned by machines.} The closest precedent is the neural-network renormalisation-group programme of \cite{Hu:2019nea}. They train a hierarchical flow-based network to realise an exact holographic mapping between uncorrelated bulk variables and a boundary field ensemble, recovering a hyperbolic emergent bulk geometry, building on the flow-based variational RG of \cite{Li:2018rnq}. Discrete neural-network formulations in which network depth represents the holographic radial direction have been developed in the AdS/deep-learning programme \cite{Hashimoto:2018ftp,PhysRevD.99.106017}, and a continuous neural ODE formulation, in which the bulk metric is learned from data, was developed in \cite{Hashimoto:2020jug}. In all of these, the holographic map is the output of learning, whereas in GenAdS it is the input, fixed analytically by the Klein--Gordon propagator. Furthermore, we are exploring what this map contributes to a standard generative task.

\vspace{\baselineskip}\noindent\emph{Physical equations as generative flows.} GenPhys \cite{Liu:2023iee} converts a broad class of partial differential equations directly into probability flows, and Poisson flow generative models \cite{Xu:2022pfgm} do the same for electrostatics in an augmented dimension. GenAdS differs structurally. The physics does not define the generative flow by itself, but is rather just an additive backbone whose deviation from the flow-matching target is learned by a residual network, in the spirit of physics-plus-learned-component decompositions such as APHYNITY \cite{Yin:2020aph}.

Physics-constrained flow matching \cite{Baldan:2025hgf} is a related direction in which the physics instead enters as a training-time constraint on the generated samples rather than as part of the velocity field. Phase-space augmentation of generative dynamics, in which one generates in a doubled space of positions and momenta, appears in critically damped Langevin diffusion \cite{Dockhorn:2021cld} and subsequent stochastic-bridge constructions \cite{Chen:2023psb}. Our work makes use of the field-theoretic counterpart to this approach, with the momentum fixed by the radial derivative of the propagator rather than sampled independently.

\vspace{\baselineskip}\noindent\emph{RG-structured and spectrally ordered generation.} Coarse-to-fine generation ordered by spatial frequency has been implemented through inverse heat dissipation \cite{Rissanen:2022ihd}, and RG-inspired multiscale generative flows have been explored in \cite{Li:2018rnq,Sheshmani:2024rgf}. The AdS encoding realises the same coarse-to-fine ordering, but the schedule and filter shape are fixed by the geometry (Section \ref{subsec:holoEncod}) rather than chosen. In Section \ref{subsec:controls}, we compare this AdS-fixed filter against generic alternatives directly.

\vspace{\baselineskip}\noindent\emph{Interpolation ans\"atze in flow matching.} Spline interpolants between marginals have been used as flow-matching paths \cite{Lee:2025hfm}. In GenAdS, the cubic Hermite structure is not an interpolation ansatz but is dictated by the on-shell constraint $\partial_r\tilde{\Phi} = \tilde{\Pi}$ of the first-order Klein--Gordon system (Section \ref{sec:flowMatching}).

\paragraph{Outline} The paper is organised as follows. Section \ref{sec:kleinGordon} develops Klein--Gordon theory in AdS/CFT, including the bulk-to-boundary propagator and its spectral decomposition. Section \ref{sec:flowMatching} then reviews flow matching and introduces the modifications, in particular to the path and the loss, that allow physics to enter. Section \ref{sec:design} combines these ingredients into the GenAdS construction, with the holographic encoding at its centre and a convolutional architecture for the velocity network, along with the implementation-level lift-noise term whose effect is quantified in our experiments. Sections \ref{sec:experimentsCheckerboard} and \ref{sec:experimentMNIST} present the experiments on the checkerboard and on MNIST, respectively. The checkerboard experiments include comparisons against matched spectral and generic-filter controls, the lift-noise ablation, and multi-seed $\Delta$ and hyperscaling-violating scans. The MNIST experiments are quantified by kernel and precision-recall metrics.

\section{Klein--Gordon theory in A\MakeLowercase{d}S/CFT}\label{sec:kleinGordon}

We first briefly review scalar field theory in AdS. The material in this section is standard knowledge in the AdS/CFT literature; see \cite{Aharony:1999ti,Skenderis:2002wp} for reviews. The purposes of this discussion are to fix conventions and to make the paper self-contained. The new construction begins in Section \ref{sec:design}. 

In general relativity, scalar field theory on AdS is most generically described by a complicated set of coupled partial differential equations. Two simplifications make the analysis more tractable. The first is to suppress gravitational backreaction, so that the AdS background is fixed. We restrict to maximally symmetric boundaries, so the bulk metric takes the following warped form:
\begin{equation}
ds^2 = dr^2 +  f(r)^2\,\widehat{g}_{ab}\,dx^a dx^b.\label{warpedAnsatz}
\end{equation}
Here, $ds^2$ is the line element, $r$ is the radial coordinate, $f$ is the warp factor, and $\widehat{g}_{ab}$ is a $d$-dimensional metric on the boundary, with indices $a,b$ running from $1$ to $d$. The radial dependence of the transverse slices is captured entirely by $f$, which means $\partial_r \widehat{g}_{ab} = 0$. Three choices of $\widehat{g}_{ab}$ correspond to the standard slicings, namely flat ($f(r) = e^r$), spherical ($f(r) = \sinh r$), and hyperbolic ($f(r) = \cosh r$). We work in Euclidean signature throughout this paper, so $\widehat{g}_{ab}$ is positive definite.

For a generic positive-definite metric $g$, the Klein--Gordon equations for a scalar field $\Phi$ are
\begin{equation}
\left(\Delta_g - m^2\right)\Phi = 0,\ \ \ \ \Delta_g \equiv \frac{1}{\sqrt{g}}\partial_\mu \left(\sqrt{g} g^{\mu\nu} \partial_\nu\right).
\end{equation}
$\Delta_g$ is the Laplace--Beltrami operator (or Laplacian) on the metric $g_{\mu\nu}$. For the warped ansatz \eqref{warpedAnsatz}, this operator decomposes into partial derivatives of $r$ and the Laplacian $\widehat{\Delta}_{g}$ on $\widehat{g}_{ab}$, so the Klein--Gordon equation takes the following form:
\begin{equation}
\left[\partial_r^2 + d\,\frac{f'(r)}{f(r)} \partial_r + \frac{1}{f(r)^2} \widehat{\Delta}_{g} - m^2\right]\Phi = 0.\label{kleinGordonWarped}
\end{equation}
It is convenient to trade $m^2$ for the scaling dimension $\Delta$ of the dual boundary operator. The holographic dictionary \cite{Aharony:1999ti}, central to AdS/CFT, relates bulk and boundary quantities; for a scalar field $\Phi$ in AdS, the dual is a scalar operator $O$ in the CFT, and the masses and scaling dimensions are connected by
\begin{equation}
m^2 = \Delta(\Delta-d).
\end{equation}
We restrict to $\Delta > d/2$, the so-called standard quantisation, in which the near-boundary modes of the scalar correspond unambiguously to boundary data (cf. \cite{Klebanov:1999tb}). This does not exclude negative $m^2$; such scalars remain stable if they satisfy the Breitenlohner--Freedman bound $m^2 > -d^2/4$ \cite{Breitenlohner:1982bm}.

We will work with the planar slicing of AdS. The spherical and hyperbolic cases are equally valid backgrounds, but the planar choice is the most tractable and is sufficient as a first-pass demonstration of GenAdS. We leave the others to future work.

\subsection{Solving Klein--Gordon with propagators}

To use Klein--Gordon dynamics inside a generative flow, we need to map boundary data to bulk fields. The standard tool is the bulk-to-boundary propagator, which we describe below.

The inverse problem, extracting CFT information from AdS fields, is usually handled with the extrapolate dictionary \cite{Banks:1998dd,Harlow:2011ke}; under the warped ansatz \eqref{warpedAnsatz} and in standard quantisation, a Klein--Gordon scalar $\Phi$ satisfies
\begin{equation}
\Phi(r,\vec{x}) \sim e^{-(d-\Delta)r} J(\vec{x}) + e^{-\Delta r} \frac{\expval{O(\vec{x})}}{2\Delta - d},
\label{asymptoticPhi}
\end{equation}
near the large-$r$ boundary, where $\vec{x}$ represents the transverse coordinates, $J$ is a CFT source, and $\expval{O}$ is the vacuum expectation value of the dual scalar operator $O$.

Only the leading source and normalisable terms are displayed in \eqref{asymptoticPhi}; the full expansion contains subleading local terms in each series and logarithms when $2\Delta - d$ is an even integer \cite{Skenderis:2002wp}. Although \eqref{asymptoticPhi} is only an asymptotic statement, it is enough to serve as a boundary condition on $\Phi$. The Klein--Gordon equation can then be solved formally in terms of a bulk-to-boundary propagator $K(r,\vec{x};\vec{x}')$, where $(r,\vec{x})$ is the bulk insertion point and $\vec{x}'$ is the boundary one. The propagator is fixed by two conditions: it solves the Klein--Gordon equation, and it exhibits $\delta$-function behaviour in the appropriate boundary limit, in standard quantisation:
\begin{equation}
\begin{split}
\left(\Delta_g - m^2\right)K(r,\vec{x};\vec{x}') &= 0,\\
\lim_{r \to \infty} e^{(d-\Delta)r}K(r,\vec{x};\vec{x}') &\propto \delta(\vec{x}-\vec{x}').
\end{split}
\end{equation}
These two conditions imply that the convolution of $K$ against any source $J$ solves the field equation with the asymptotic behaviour \eqref{asymptoticPhi}, up to an overall normalisation. We accordingly write
\begin{equation}
\Phi(r,\vec{x}) \equiv \int d^d x'\sqrt{\widehat{g}}\,K(r,\vec{x};\vec{x}')J(\vec{x}').\label{fieldSoln}
\end{equation}
The propagator itself can be written explicitly. Defining the normalisation coefficient,
\begin{equation}
C_\Delta \equiv \frac{\Gamma(\Delta)}{\pi^{d/2}\Gamma(\Delta - d/2)},
\end{equation}
the planar bulk-to-boundary propagator is
\begin{equation}
K(r,\vec{x};\vec{x}') = \frac{C_\Delta}{\big(e^r |\vec{x}-\vec{x}'|^2 + e^{-r}\big)^{\Delta}},\label{propPlanar}
\end{equation}
where $|\vec{x}-\vec{x}'|$ is the Euclidean distance between $\vec{x}$ and $\vec{x}'$. The normalisation is chosen to make \eqref{fieldSoln} agree with \eqref{asymptoticPhi}.

\subsection{Spectral decomposition of Klein--Gordon}

Flow matching is built on first-order ODEs \cite{chen2019neuralordinarydifferentialequations,lipman2023flowmatchinggenerativemodeling}, whereas the Klein--Gordon equation is a partial differential equation. To bridge this gap, we recast Klein--Gordon as a family of ODEs by spectral decomposition. The starting point is the eigenproblem for the transverse Laplacian $\widehat{\Delta}_g$,
\begin{equation}
\widehat{\Delta}_g Y_{\lambda}^{\alpha}(\vec{x}) = -\lambda Y_{\lambda}^{\alpha}(\vec{x}).
\end{equation}
$\lambda$ labels the spectral modes (discrete for bounded transverse spaces, continuous otherwise), and $\alpha$ is the degeneracy label. We impose orthonormality,
\begin{equation}
\int d^d x \sqrt{\widehat{g}}\, Y_\lambda^{\alpha}(\vec{x}) \bar{Y}_{\lambda'}^{\alpha'}(\vec{x}) = \delta(\lambda-\lambda') \delta_{\alpha\alpha'},\label{orthonormality}
\end{equation}using bars to represent complex conjugation. The $\delta(\lambda - \lambda')$ here is taken with respect to a generic measure $d\mu_\lambda$, in the sense that $\int d\mu_\lambda\, f(\lambda) \delta(\lambda - \lambda') = f(\lambda')$. \eqref{orthonormality} implies a completeness relation,
\begin{equation}
\delta(\vec{x}-\vec{x}') = \int d\mu_\lambda \sum_\alpha Y_{\lambda}^{\alpha}(\vec{x}) \bar{Y}_{\lambda}^{\alpha}(\vec{x}').
\end{equation}
Applying the spectral decomposition to Klein--Gordon yields,
\begin{equation}
\Phi(r,\vec{x}) = \int d\mu_\lambda\,\sum_\alpha \phi_{\lambda\alpha}(r)\, Y_{\lambda}^\alpha(\vec{x}),
\end{equation}
where a generic scalar carries $\alpha$-dependent mode coefficients $\phi_{\lambda\alpha}$. Since the transverse Laplacian is blind to $\alpha$, each $\phi_{\lambda\alpha}$ satisfies the same radial equation. The Klein--Gordon equation \eqref{kleinGordonWarped} reduces to a one-dimensional ODE for these coefficients,
\begin{equation}
\frac{d^2 \phi_\lambda}{dr^2} + d\,\frac{f'(r)}{f(r)} \frac{d\phi_\lambda}{dr} = \left[\frac{\lambda}{f(r)^2} + \Delta(\Delta-d)\right]\phi_\lambda(r),\label{ODEmodesGen}
\end{equation}
where we henceforth suppress the $\alpha$ subscripts of the coefficients. This is harmless in the present discussion because of our assumption of maximal symmetry of the transverse slices. The bulk-to-boundary propagator admits an analogous expansion, depending on both $\vec{x}$ and $\vec{x}'$,
\begin{equation}
K(r,\vec{x};\vec{x}') = \int d\mu_\lambda\,\kappa_\lambda(r) \sum_\alpha Y_\lambda^\alpha(\vec{x}) \bar{Y}_\lambda^\alpha(\vec{x}').
\end{equation}
The coefficients $\kappa_\lambda$ (with $\alpha$-dependence again suppressed) satisfy the same ODE \eqref{ODEmodesGen}, and they are related to $\phi_\lambda$ through the source coefficients $j_{\lambda}$,
\begin{equation}
\phi_\lambda(r) = j_\lambda \kappa_\lambda(r),\ \ J(\vec{x}) = \int d\mu_\lambda\,j_{\lambda}\sum_{\alpha} Y_\lambda^\alpha(\vec{x}).\label{fourierConvolve}
\end{equation}
For the planar slicing, the transverse space is unbounded, so the spectrum is continuous. Plane waves diagonalise the Laplacian,
\begin{equation}
\widehat{\Delta}_g e^{-i \vec{k} \cdot \vec{x}} = -|\vec{k}|^2 e^{-i \vec{k} \cdot \vec{x}},
\end{equation}
so the eigenbasis is a Fourier basis labelled by momentum $\vec{k} = (k_1,\dots,k_d)$, with the spectral parameter equal to the squared norm $|\vec{k}|^2$. The integration measure over modes is the flat one $d^d k$. The normalisation
\begin{equation}
\int d^d k\,e^{-i \vec{k} \cdot \vec{x}} e^{i \vec{k} \cdot \vec{x}'} = (2\pi)^d \delta(\vec{x}-\vec{x}')
\end{equation}
fixes the basis as
\begin{equation}
Y_{\vec{k}}(\vec{x}) \equiv \frac{1}{(2\pi)^{d/2}}e^{-i \vec{k} \cdot \vec{x}}.
\end{equation}
A Fourier transform of \eqref{propPlanar} now gives the propagator coefficients,
\begin{equation}
\kappa_{|\vec{k}|}(r)
= \int d^d x\,e^{i\vec{k} \cdot \vec{x}}K(r,\vec{x};0) = \frac{2}{\Gamma(\nu)} e^{-dr/2} \left(\frac{|\vec{k}|}{2}\right)^\nu K_\nu(|\vec{k}|e^{-r}),
\end{equation}
where $K_\nu$ is the modified Bessel function of the second kind and $\nu \equiv \Delta - d/2$.

\subsection{First-order formulation}

To use Klein--Gordon dynamics inside flow matching, we need to rewrite it in its first-order formulation. We introduce the canonical momentum,
\begin{equation}
\Pi \equiv \partial_r \Phi,
\end{equation}
with which we recast Klein--Gordon as Hamilton's equations,
\begin{align}
\partial_r \Phi &= \Pi,\label{kg1}\\
\partial_r \Pi &= \left[\Delta(\Delta-d) - \dfrac{1}{f(r)^2}\widehat{\Delta}_g\right]\Phi - d\,\dfrac{f'(r)}{f(r)}\Pi.\label{kg2}
\end{align}
In spectral form and with mode coefficients $\phi_\lambda$ and $\pi_\lambda$, the system becomes
\begin{align}
\frac{d\phi_\lambda}{dr} &= \pi_\lambda(r),\label{kg1spectral}\\
\frac{d\pi_\lambda}{dr} &= \left[\dfrac{\lambda}{f(r)^2} + \Delta(\Delta-d)\right]\phi_\lambda(r) - d\,\dfrac{f'(r)}{f(r)}\pi_\lambda(r).\label{kg2spectral}
\end{align}
Equations \eqref{kg1spectral}--\eqref{kg2spectral} define a trajectory in phase space, and they will supply inductive bias for the flow trajectory.

\subsection{Numerical stabilisation of fields}\label{subsec:radRescale}

The asymptotic form \eqref{asymptoticPhi} causes a numerical problem. The source term scales as $e^{-(d-\Delta)r}$, so it decays toward the boundary for $\Delta < d$, is $r$-independent for $\Delta = d$, and grows for $\Delta > d$. In every case, it must be cleanly separated from the VEV term, particularly in the decaying case in which the source becomes exponentially suppressed precisely where the model needs to extract it. The remedy is a field redefinition,
\begin{equation}
\tilde{\Phi} \equiv e^{(d-\Delta) r}\Phi,\ \ \ \ \tilde{\Pi} \equiv e^{(d-\Delta) r}\left[\Pi + (d-\Delta) \Phi\right].\label{fieldRedefs}
\end{equation}
Under this redefinition, the VEV term in \eqref{asymptoticPhi} decays exponentially for any $\Delta > d/2$ while the source term remains $O(1)$, so the latter can be cleanly isolated.

The redefined fields satisfy a slightly modified version of the first-order Klein--Gordon system. In planar AdS and in spectral form, with coefficients $\tilde{\phi}_{\vec{k}}$ and $\tilde{\pi}_{\vec{k}}$,
\begin{align}
\frac{d\tilde{\phi}_{\vec{k}}}{dr} &= \tilde{\pi}_{\vec{k}}(r),\label{kg1stabl}\\
\frac{d\tilde{\pi}_{\vec{k}}}{dr} &=  |\vec{k}|^2 e^{-2r} \tilde{\phi}_{\vec{k}}(r) - \left(2\Delta-d\right)\tilde{\pi}_{\vec{k}}(r).\label{kg2stabl}
\end{align}
The propagator is rescaled by the same factor as the field,
\begin{equation}
\tilde{K} \equiv e^{(d-\Delta) r}K,
\end{equation}
which simply shifts its mode coefficients by $e^{(d-\Delta)r}$,
\begin{equation}
\tilde{\kappa}_{|\vec{k}|}(r) = \frac{2}{\Gamma(\nu)} \left(\frac{|\vec{k}|e^{-r}}{2}\right)^\nu K_\nu(|\vec{k}|e^{-r}).\label{kappaTilde}
\end{equation}

\section{Elements of flow matching}\label{sec:flowMatching}

We now turn to the machine-learning side of the discussion, reviewing the flow-matching framework of \cite{lipman2023flowmatchinggenerativemodeling} and the modifications that allow Klein--Gordon to enter as inductive bias.

We use $S \equiv (\tilde{\Phi},\tilde{\Pi})$ to represent a point in redefined phase space \eqref{fieldRedefs}, and write components with superscripts, i.e. $\Psi = (\Psi^{\tilde{\Phi}},\Psi^{\tilde{\Pi}})$. The first-order Klein--Gordon equations condense to $\partial_r S = V_{\text{KG}}$, with
\begin{equation}
V_{\text{KG}} = \left(\tilde{\Pi},-e^{-2r}\widehat{\Delta}_g \tilde{\Phi} - (2\Delta-d)\tilde{\Pi}\right).
\end{equation}
The flow lives in this phase space and is parameterised by the radial coordinate of AdS. For numerical purposes, we truncate at $r = r_{\text{IR}}$ (deep in the bulk) and $r = r_{\text{UV}}$ (near the boundary), and we introduce a compact time coordinate $t \in [0,1]$ by
\begin{equation}
r(t) = (r_{\text{UV}} - r_{\text{IR}})t + r_{\text{IR}}.\label{tparam}
\end{equation}
This $t$ is not physical time; it is a reparameterisation of the AdS radial coordinate. The flow is then $S_t$, with $S_0$ sampled from a base distribution (typically a Gaussian) and $S_1$ from the target.

The job of flow matching is to train a network to learn the velocity field $V_t$. Once trained, generation proceeds by sampling the base distribution and integrating numerically.

\subsection{Flows and log-likelihood computation}

A flow-based generative model is governed by an ODE,
\begin{equation}
\frac{dS_t}{dt} = V_t(\theta),
\end{equation}
where the velocity $V_t(\theta)$ is a neural network with parameters $\theta$. Such an equation is called a neural ODE \cite{chen2019neuralordinarydifferentialequations}. In a continuous normalising flow (CNF), one transports a probability density along this velocity field. Conservation of probability gives a continuity equation for this flow,
\begin{equation}
\pdv{p_t}{t} = -\nabla_a \left(p_t V_t^a\right),
\end{equation}
where $a$ indexes the dimensions of the flow space and $p_t$ is the density at time $t$. Our flows live in flat phase space, so the continuity equation can be integrated to give the log likelihood,
\begin{equation}
\log p_1 = \log p_0 - \int_0^1 dt\left(\pdv{V_t^{\tilde{\Phi}}}{\tilde{\Phi}} + \pdv{V_t^{\tilde{\Pi}}}{\tilde{\Pi}}\right).\label{logLike}
\end{equation}
Direct CNF training \cite{chen2019neuralordinarydifferentialequations,grathwohl2018ffjordfreeformcontinuousdynamics} uses this log likelihood as a loss, but evaluating it requires numerical integration at every step, which is expensive.

\subsection{Linear versus Hermite paths}

Flow matching is cheaper to train than a CNF because the path between base and target is fixed in advance, eliminating the need to integrate during training. The vanilla choice \cite{lipman2023flowmatchinggenerativemodeling} is a linear path,
\begin{equation}
S_t = (1-t)S_0 + t S_1,\label{linearPathDef}
\end{equation}
with $S_0$ drawn from the base distribution and $S_1$ from the target. The corresponding velocity is constant in time, $\partial_t S_t = S_1 - S_0$, and the network is trained to reproduce it.

This linear path is simple but it is off-shell, since it does not respect the first Klein--Gordon equation $\partial_r \tilde{\Phi} = \tilde{\Pi}$. To see this, one must first fix an interpretation of the endpoints. We treat them as fixed phase-space samples; the pairs $(\tilde{\Phi}_0,\tilde{\Pi}_0)$ and $(\tilde{\Phi}_1,\tilde{\Pi}_1)$ are drawn once, at $r_{\text{IR}}$ and $r_{\text{UV}}$ respectively, and carry no $r$-dependence. Since $\tilde{\Phi}_t$ is linear in $t$ and $r(t)$ is affine, differentiating along the path and comparing against the interpolated momentum $\tilde{\Pi}_t = (1-t)\tilde{\Pi}_0 + t\tilde{\Pi}_1$ gives
\begin{equation}
\partial_r \tilde{\Phi}_t - \tilde{\Pi}_t = \frac{\tilde{\Phi}_1 - \tilde{\Phi}_0}{r_{\text{UV}} - r_{\text{IR}}} - \left[(1-t)\tilde{\Pi}_0 + t\tilde{\Pi}_1\right],
\end{equation}
which does not vanish for generic endpoint samples. In contrast, an on-shell constraint would require $\partial_r \tilde{\Phi}_t = \tilde{\Pi}_t$, which the linear path violates.

The fix is to interpolate $\tilde{\Phi}_t$ alone, then define $\tilde{\Pi}_t$ as $\partial_r \tilde{\Phi}_t$. We have four endpoint constraints, two on $\tilde{\Phi}$ and two on $\tilde{\Pi}$, so the minimal smooth choice is a cubic. Spline interpolants between marginals have also been used in flow matching \cite{Lee:2025hfm}. We write it in the cubic Hermite basis, whose four polynomials on $[0,1]$ are
\begin{equation}
\begin{array}{ll}
H_{00}(u) = 2u^3 - 3u^2 + 1,\ &H_{01}(u) = -2u^3 + 3u^2,\\
H_{10}(u) = u^3 - 2u^2 + u,\ &H_{11}(u) = u^3 - u^2.
\end{array}
\end{equation}
The subscripts on $H_{ij}$ have a systematic meaning. The first index $i \in \{0,1\}$ indicates whether the basis function controls a value ($i=0$) or a derivative ($i=1$) at an endpoint, while the second index $j \in \{0,1\}$ indicates which endpoint is meant ($j=0$ for $u=0$, $j=1$ for $u=1$). Numerically, $H_{00}(0) = H_{01}(1) = 1$ for the two value-basis functions, and $H'_{10}(0) = H'_{11}(1) = 1$ for the two derivative-basis functions.

A path parameter $u \in [0,1]$ designed to absorb the changing volume of the radial slices of AdS as $r$ runs from $r_{\text{IR}}$ to $r_{\text{UV}}$ proves convenient. We define it as
\begin{equation}
u(r) = \left[\int_{r_{\text{IR}}}^{r_{\text{UV}}} \frac{dr'}{\sqrt{g(r')}}\right]^{-1} \int_{r_{\text{IR}}}^{r} \frac{dr'}{\sqrt{g(r')}},\label{affineParams}
\end{equation}
which for planar AdS evaluates to
\begin{equation}
u(r) = \frac{1-e^{-d(r - r_{\text{IR}})}}{1-e^{-d(r_{\text{UV}} - r_{\text{IR}})}}.
\end{equation}
The path is then
\begin{equation}
\tilde{\Phi}_u = H_{00}(u) \tilde{\Phi}_0 + H_{01}(u)\tilde{\Phi}_1 + H_{10}(u) m_0 + H_{11}(u) m_1,
\end{equation}
with
\begin{equation}
m_0 =  \left.\left(\frac{dr}{du}\right)\right|_{r_{\text{IR}}}\tilde{\Pi}_0,\ \ \ \ m_1 = \left.\left(\frac{dr}{du}\right)\right|_{r_{\text{UV}}}\tilde{\Pi}_1.
\end{equation}
$\tilde{\Phi}_t$ follows by composing $u \to r \to t$, and $\tilde{\Pi}_t$ from $\partial_r \tilde{\Phi}_t$.

\subsection{Losses and residuals}

With both paths in hand, we can write the losses. It is convenient to define
\begin{equation}
\delta_r \equiv \frac{dr}{dt} = r_{\text{UV}} - r_{\text{IR}},
\end{equation}
since the velocities entering the loss are $t$-derivatives. All experiments in this paper use $\delta_r = 1$.

For the linear path, the velocities are constant along the path,
\begin{align}
U_t^{\tilde{\Phi}} = \tilde{\Phi}_1 - \tilde{\Phi}_0,\ \ \ \ U_t^{\tilde{\Pi}} = \tilde{\Pi}_1 - \tilde{\Pi}_0.\label{linearPath}
\end{align}
In contrast, for the Hermite path, the velocities are
\begin{equation}
U_t^{\tilde{\Phi}} = \delta_r \tilde{\Pi}_t,\ \ \ \ U_t^{\tilde{\Pi}} = \delta_r\left[\left(\frac{du}{dr}\right)^2 \partial_u^2 \tilde{\Phi}_t + \frac{d^2 u}{dr^2} \partial_u \tilde{\Phi}_t\right].\label{hermitePath}
\end{equation}
A network $V_t(\theta)$ is trained to learn these velocities. A naive mean-squared-error loss would ignore the warped metric \eqref{warpedAnsatz}; we use instead a warped loss,
\begin{equation}
\mathcal{L} = \expval{\left|\left|V_t(\theta) - U_t\right|\right|_{g_{r(t)}}^2}_{t},\label{fullLoss}
\end{equation}
where the average $\expval{\cdot}$ runs over $t$ as well as over the endpoint samples, and the warped norm of a phase-space function $\Psi(S)$ is
\begin{equation}
\left|\left|\Psi(S)\right|\right|^2_{g_{r(t)}}
= \int d^d x\,f(r)^d \left[|\Psi^{\tilde{\Phi}}(r,\vec{x})|^2 + |\Psi^{\tilde{\Pi}}(r,\vec{x})|^2\right],
\end{equation}
i.e. a sum weighted by the volumes of the radial slices.

So far, the loss does not incorporate Klein--Gordon explicitly. To bring it in, we introduce a residual network $R_t(\theta)$ and substitute
\begin{equation}
V_t(\theta) \to \delta_r V_{\text{KG}} + R_t(\theta).
\end{equation}
The Klein--Gordon equation now serves as a backbone velocity, and $R_t(\theta)$ supplies the correction learnt by the network. The residual loss is
\begin{equation}
\mathcal{L} = \expval{\left|\left|R_t(\theta) - \left(U_t - \delta_r V_{\text{KG}}\right)\right|\right|_{g_{r(t)}}^2}_t.\label{residLoss}
\end{equation}
We now have two binary design choices, namely whether to include the Klein--Gordon backbone and which path to use. Both paths are compatible with the residual loss. The Hermite path, however, satisfies the first Klein--Gordon equation by construction, so $R_t^{\tilde{\Phi}}$ would be driven to zero in training; we therefore set $R_t^{\tilde{\Phi}}(\theta) = 0$ and train only the $\tilde{\Pi}$ component in the Hermite case.

\section{Designing the model}\label{sec:design}

The two ingredients are now in place, an AdS Klein--Gordon theory with a bulk-to-boundary propagator, and a flow-matching framework that admits physical equations of motion as inductive bias. To put them together we still need a way to convert raw data into bulk fields. We do this through a holographic encoding, work in Fourier space so the resulting flows are ODEs, and choose the velocity network to be convolutional, so as to respect the appropriate symmetries.

\subsection{Holographic encoding}\label{subsec:holoEncod}

The holographic encoding treats each ground-truth sample as a CFT source and lifts it into the bulk via the propagator. Concretely, given a sample, we identify it with a boundary source $J_*(\vec{x})$ and convolve as in \eqref{fieldSoln},
producing the following encoded scalar:
\begin{equation}
\Phi(r,\vec{x}|J_*) = \int d^d \vec{x}'\,K(r,\vec{x};\vec{x}')J_*(\vec{x}').\label{fieldEncode}
\end{equation}
The boundary value $\Phi(r_{\text{UV}},\vec{x}|J_*)$ then plays the role of training data. We emphasise that $K$ is the bulk-to-boundary propagator of a specific physical theory (e.g. Klein--Gordon as in Section \ref{sec:kleinGordon}), so this is a highly specific feature embedding of the boundary data $J_*$ into AdS space.

The bulk scalar being a free probe on a fixed background yields two properties of this embedding. First, the field is linear in the source. As such, the dual boundary field is a generalised free field \cite{Aharony:1999ti,Duetsch:2002gd}, so the analytic holographic prior supplied by the field is limited to two-point (Gaussian) structure. All non-Gaussian structure required to model the empirical source distribution must therefore be learned by the residual network of Section \ref{sec:flowMatching}, rather than supplied by the free-field backbone; the controls of Section \ref{subsec:controls} quantify what remains attributable to the encoding itself. Second, by the holographic-RG identification of the radial coordinate \cite{deBoer:1999tgo,Verlinde:1999xm,Heemskerk:2010hk,Faulkner:2010jy}, the convolution \eqref{fieldEncode} is not an arbitrary feature map but admits a natural interpretation as a scale-dependent representation of the source $J_*$ at radial scale $r$. This interpretation is quantified by an induced scale-dependent filter, which we discuss below.

For 2-dimensional point data we use a point encoding. Each sample is a point $\vec{x}_* \in \mathbb{R}^2$, mapped to a $\delta$ source on the boundary,
\begin{equation}
J_*(\vec{x}) = \delta(\vec{x}-\vec{x}_*),\label{deltaSource}
\end{equation}
so that \eqref{fieldEncode} reduces to evaluating the propagator at $\vec{x}_*$,
\begin{equation}
\Phi(r_{\text{UV}},\vec{x}|J_*) = K(r_{\text{UV}},\vec{x};\vec{x}_*).
\end{equation}
For an image, $J_*$ is taken to be the pixel intensity map, the scalar-valued function on the pixel grid giving each pixel's greyscale intensity. The integral \eqref{fieldEncode} is no longer evaluable in closed form, but it becomes tractable in Fourier space, where the source can be spectrally decomposed.

\begin{figure}[!htbp]
\centering
\includegraphics{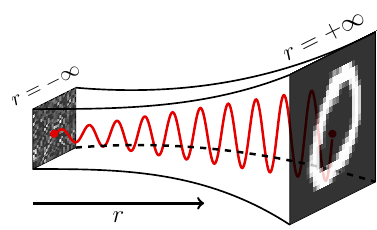}
\caption{A schematic representation of the holographic encoding for a sample of the MNIST dataset. The image is treated as a source on the boundary for the corresponding bulk field, which flows to noise.}
\label{figs:holographicEncoding}
\end{figure}

To train the model, the network learns the velocity of a flow that runs from a noise profile deep in the bulk to the holographic encoding of a sample near the boundary. Figure \ref{figs:holographicEncoding} shows this schematically.

\subsection{Spectral representation of data}

Flow matching needs ODEs, not PDEs, so the machinery must be translated into Fourier space. This also makes the holographic encoding of images tractable, since in Fourier space the convolution \eqref{fieldEncode} becomes a multiplication. For a Fourier mode of momentum $\vec{k}$, the corresponding mode coefficients of the encoded field follow from \eqref{fourierConvolve},
\begin{equation}
\begin{split}
\phi_{\vec{k}}(r_{\text{UV}}|J_*) &= j_{*\vec{k}}\kappa_{|\vec{k}|}(r_{\text{UV}}),\\
\pi_{\vec{k}}(r_{\text{UV}}|J_*) &= j_{*\vec{k}}\partial_r \kappa_{|\vec{k}|}(r_{\text{UV}}),
\end{split}
\end{equation}
where $j_{*\vec{k}}$ is the source mode coefficient,
\begin{equation}
j_{*\vec{k}} = \int \frac{d^d x}{(2\pi)^{d/2}} e^{i\vec{k} \cdot \vec{x}} J_*(\vec{x}).
\end{equation}
For the $\delta$ source \eqref{deltaSource}, the source mode coefficients are simply $e^{i \vec{k} \cdot \vec{x}_*}/(2\pi)^{d/2}$, giving
\begin{equation}
\begin{split}
\phi_{\vec{k}}(r_{\text{UV}}|\vec{x}_*) &= \frac{1}{(2\pi)^{d/2}} \kappa_{|\vec{k}|}(r_{\text{UV}}) e^{i\vec{k} \cdot \vec{x}_*},\\
\pi_{\vec{k}}(r_{\text{UV}}|\vec{x}_*) &= \frac{1}{(2\pi)^{d/2}} \partial_r \kappa_{|\vec{k}|}(r_{\text{UV}}) e^{i\vec{k} \cdot \vec{x}_*}.
\end{split}
\end{equation}
In this spectral point encoding, the position $\vec{x}_*$ shows up as a phase factor. In practice the code uses the analogous expressions for the redefined fields of Section \ref{subsec:radRescale}. From \eqref{kappaTilde},
\begin{align}\label{trueSamples}
\tilde{\phi}_{\vec{k}}(r_{\text{UV}}|J_*) &= \frac{2j_{*\vec{k}}}{\Gamma(\nu)} \left(\frac{|\vec{k}| e^{-r_{\text{UV}}}}{2}\right)^\nu K_\nu(|\vec{k}|e^{-r_{\text{UV}}}),\\
\tilde{\pi}_{\vec{k}}(r_{\text{UV}}|J_*) &= \frac{4 j_{*\vec{k}}}{\Gamma(\nu)} \left(\frac{|\vec{k}| e^{-r_{\text{UV}}}}{2}\right)^{\nu + 1} K_{\nu-1}(|\vec{k}|e^{-r_{\text{UV}}}).\nonumber
\end{align}
In this convention, the source mode $j_{*\vec{k}}$ is multiplied by a Bessel factor with argument $|\vec{k}|e^{-r}$. This factor is a scale-dependent low-pass filter. Near the UV boundary ($|\vec{k}|e^{-r}\ll 1$) the asymptotic behaviour $K_\nu(x)\sim (2/x)^\nu \Gamma(\nu)/2$ yields $\tilde{\kappa}_{|\vec{k}|}(r)\to 1$, so all modes pass through. Meanwhile deeper in the bulk ($|\vec{k}|e^{-r}\gg 1$), the asymptotic behaviour $K_\nu(x)\sim\sqrt{\pi/(2x)}e^{-x}$ implies an exponential decay $\tilde{\kappa}_{|\vec{k}|}(r)\sim e^{-|\vec{k}|e^{-r}}$ that cuts off modes with $|\vec{k}|\gtrsim e^r$. As such, the flow from IR to UV is intrinsically multiscale; Fourier modes of increasingly high wavenumber enter as $r$ grows, inducing a coarse-to-fine ordering set by the AdS geometry rather than by architecture. This is the quantitative form of the RG reading above \cite{deBoer:1999tgo,Verlinde:1999xm,Heemskerk:2010hk,Faulkner:2010jy}, and in the present encoding, the radial coordinate sets the characteristic resolution scale of the source representation. Connections between RG flow, optimal transport, and diffusion-based generative modelling are developed in \cite{Cotler:2022fze,Cotler:2023lem}.

\paragraph{Lift noise.} One detail of our implementation of the encoding must be made explicit here. The implementation supports perturbing the momentum coefficients $\tilde{\pi}_{\vec{k}}$ of each training sample by independent Gaussian noise of standard deviation $\sigma$ (parameter \texttt{lift\_noise\_sigma}) after the lift \eqref{trueSamples}. The noise term was introduced as a regularising perturbation of the momentum channel and intended to prevent the network from overfitting the exact derivative relation between $\tilde{\Phi}$ and $\tilde{\Pi}$. Physically, the perturbed UV training state no longer lies exactly on the propagator lift of the source. Consequently, the Hermite path of Section \ref{sec:flowMatching}, whose endpoint slope $m_1$ includes $\tilde{\Pi}_1$ directly, has its training targets corrupted rather than merely its inputs.

The ablation of this noise term materially affects the experimental conclusions, as we see in Section \ref{subsec:liftnoise}. As such, the default in this paper and in the public code is $\sigma = 0$, and all results below are reported at this default unless explicitly stated otherwise.

\subsection{Convolutional networks for spectral flows}

We now turn to the implementation of the flow in Fourier space. Bounding the transverse space induces a discretisation. Truncating each spatial component $x_i$ to an interval of length $L_i$ quantises the corresponding momentum component as
\begin{equation}
k_i = \frac{2\pi m}{L_i},\ \ \ \ m \in \mathbb{Z}.
\end{equation}
Taking equal lengths and keeping $K$ modes per direction, the Fourier space becomes a $d$-dimensional grid of $K^d$ points, each carrying a mode coefficient. The network learns the velocities (or residual velocities) of these mode coefficients, $d\tilde{\phi}_{\vec{k}}/dt$ and $d\tilde{\pi}_{\vec{k}}/dt$. With inputs and outputs arranged on a grid, a convolutional network is the natural choice, since it builds in translational equivariance.

One clarification is in order. The CNN acts on the momentum grid, so the symmetry it enforces is equivariance under shifts of that grid. Physical translations of the boundary act instead as momentum-dependent phases $e^{i\vec{k}\cdot\vec{a}}$ on the Fourier coefficients, which the architecture does not enforce. Grid-shift equivariance is therefore an architectural statement about weight sharing across modes, not an implementation of the physical translation symmetry.

Training is straightforward in this setup. Integrals over Fourier space reduce to discrete sums, and the warped norm $||\cdot||^2_{g_{r(t)}}$ entering the loss \eqref{fullLoss} becomes
\begin{equation}
||\Psi(S)||_{g_{r(t)}}^2 = \sum_{\vec{k}} f(r)^d \left(|\psi_{\vec{k}}^{\tilde{\Phi}}|^2 + |\psi_{\vec{k}}^{\tilde{\Pi}}|^2\right),
\end{equation}
writing the mode coefficients of $\Psi = (\Psi^{\tilde{\Phi}},\Psi^{\tilde{\Pi}})$ as $\psi_{\vec{k}}^{\tilde{\Phi}}$ and $\psi_{\vec{k}}^{\tilde{\Pi}}$.

The flow endpoints used in training have a simple form. At $t = 1$ ($r = r_{\text{UV}}$) the samples are given by \eqref{trueSamples}, and at $t = 0$ ($r = r_{\text{IR}}$) the base coefficients are drawn from Gaussians,
\begin{equation}
\begin{split}
\tilde{\phi}_{\vec{k}}(r_{\text{IR}}) &\sim \mathcal{N}\big(0,c_\phi(1+|\vec{k}|^2)^{-s_\phi}\big),\\
\tilde{\pi}_{\vec{k}}(r_{\text{IR}}) &\sim \mathcal{N}\big(0,c_\pi(1+|\vec{k}|^2)^{-s_\pi}\big).
\end{split}\label{baseDist}
\end{equation}
After training, we generate new samples by drawing from the base distribution and integrating the learned velocity numerically. The result is a set of $K^d$ Fourier mode coefficients at the boundary, which then need to be decoded.

Decoding is different for images and points. For images, the mode coefficients reconstruct a position-space source which is read as the intensity map of the generated image. For points, the generated source is a collection of $K^d$ position-space points, and we average their positions to extract a single point. The averaging is consistent with the training setup; each training sample is encoded as a $\delta$ source \eqref{deltaSource}, so a well-trained model is trained to output profiles concentrated near a single point, for which the centroid is a faithful summary. A natural refinement, which we leave to future work, is to treat the generated profile as a density on $\mathbb{R}^d$ and draw multiple points from it, thereby recovering full point-cloud statistics of each generated sample.

\section{Checkerboard experiments}\label{sec:experimentsCheckerboard}

We test GenAdS on the checkerboard distribution, the toy benchmark used in the original flow-matching paper \cite{lipman2023flowmatchinggenerativemodeling}. The checkerboard is two-dimensional and lets us isolate effects of the holographic encoding without the complications of high-dimensional data; the MNIST experiments of Section \ref{sec:experimentMNIST} stress the construction in the opposite direction.

Most hyperparameters are kept fixed across the experiments. The checkerboard sits in two dimensions, so $d = 2$. The velocity CNN is a standard stack of 2D convolutional layers acting on the $K\times K$ Fourier grid. The architectural specifications (depth, channel widths, kernel sizes, activation functions, and normalisation) are documented in the accompanying code repository, and they are such that the CNN has 10,596,868 parameters when computing the Hermite path and 10,599,176 parameters for the linear path. Training uses 50,000 samples with batch size 64, AdamW with learning rate $3 \times 10^{-4}$ and weight decay $1 \times 10^{-5}$. Generated datasets contain 10,000 points.

Several flow-matching hyperparameters are also fixed. The radial cutoffs are $r_{\text{IR}} = 0$ and $r_{\text{UV}} = 1$. The transverse plane is truncated to a square of side $L = 8$ with $K = 16$ modes per direction, giving 256 Fourier modes in total. The base distribution \eqref{baseDist} is parameterised by
\begin{equation}
c_\phi = 1.0,\ \ c_\pi = 0.55,\ \ \ \ s_\phi = 1.0,\ \ s_\pi = 1.0.\label{paramsIRDist}
\end{equation}
For each experiment we state the number of epochs and the value(s) of $\Delta$ used. Unless stated otherwise, every reported number is a mean $\pm$ one standard deviation over independent seeds, with the seed count (three or five) stated per table or figure. The lift-noise parameter discussed in Section \ref{subsec:holoEncod} is set to its default $\sigma = 0$ except where a value is given explicitly.

The checkerboard is two-dimensional, so we use the point encoding of Section \ref{subsec:holoEncod}. Ground-truth samples are then represented by the Fourier mode coefficients \eqref{trueSamples}, with
\begin{equation}
j_{* \vec{k}} = \frac{e^{i\vec{k} \cdot \vec{x}_*}}{2\pi}.
\end{equation}

\subsection{Checkerboard metrics}

A model can fail on the checkerboard in two ways: it can generate points in regions where the true checkerboard does not have any support, and it can develop non-uniformities within tiles where the true distribution is uniform. We define two metrics, one for each possibility.

The first failure mode is captured by a \emph{boundary violation} (BV) statistic, simply the fraction of generated points that fall outside valid checkerboard cells. BV measures how well the model has learned the location of the checkerboard's boundaries.

The second is captured by the \emph{within-cell energy distance} (WED). For two distributions $P$ and $Q$ with samples $\vec{x},\vec{x}' \sim P$ and $\vec{y},\vec{y}' \sim Q$, the energy distance is
\begin{equation}
\text{ED} = 2\expval{|\vec{x}-\vec{y}|}_{P,Q} - \expval{|\vec{x}-\vec{x}'|}_P - \expval{|\vec{y}-\vec{y}'|}_Q,
\end{equation}
where $|\cdot|$ is Euclidean distance. WED is the energy distance computed within each individual checkerboard tile, weighted by the number of true points per cell. Since the checkerboard is uniform within each cell, WED measures local non-uniformity of the learned distribution.

\subsection{Preliminary checkerboard ablations}\label{subsec:ablation_checkerboard}

\begin{figure}[!ht]
\centering
\includegraphics{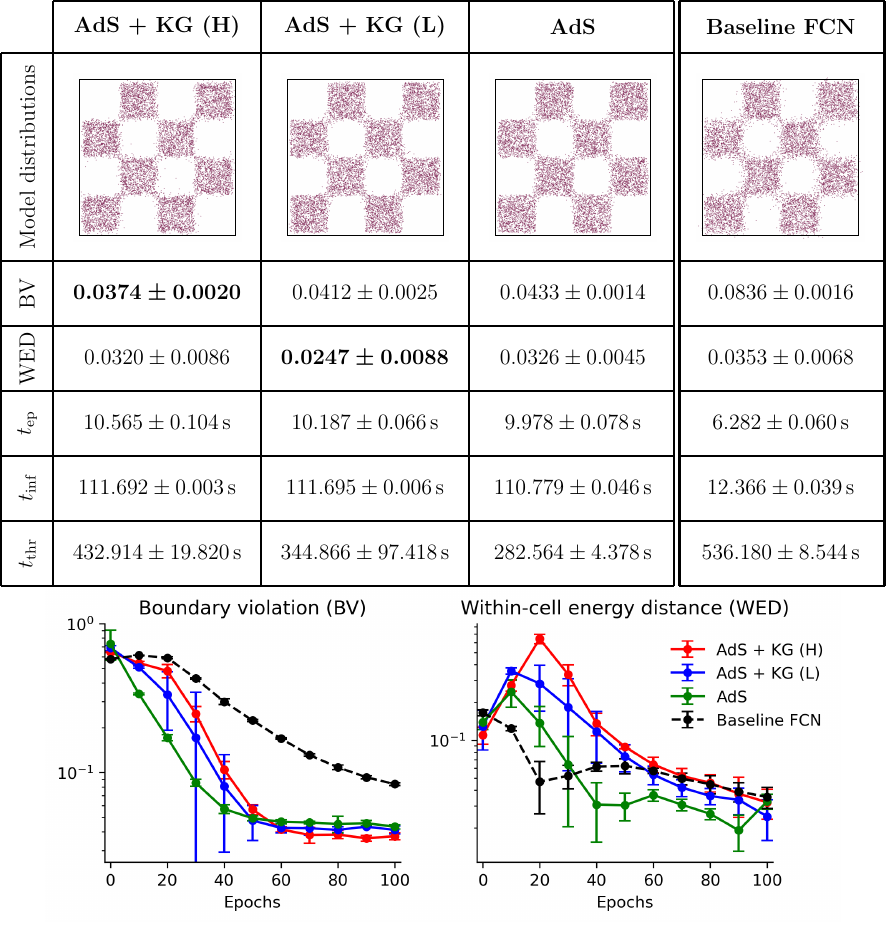}
\caption{\justifying
We present the learned distributions of models trained on the checkerboard after 100 epochs. The \textbf{AdS + KG (H)} and \textbf{AdS + KG (L)} models are trained with the loss \eqref{residLoss}, using Hermite and linear paths respectively, and the \textbf{AdS} model is trained with the loss \eqref{fullLoss} and a linear path. The \textbf{Baseline FCN} uses no AdS information. We also show the final boundary-violation (BV) and within-cell-energy-distance (WED) metrics, the time per epoch $t_{\text{ep}}$, final inference time $t_{\text{inf}}$ for 10,000 samples, and estimated threshold time $t_{\text{thr}}$ when $\text{BV} < 0.1$ (assuming linearity of BV within each 10-epoch interval), all averaged over three seeds. Additionally, we plot the BV and WED metrics. These runs use the lift-noise setting $\sigma = 0.1$ (Section \ref{subsec:holoEncod}). The corresponding comparison at the default $\sigma = 0$, including the matched controls, is given in Table \ref{tab:sigma0_checkerboard}.}
\label{fig:ablations_checkerboard}
\end{figure}

Figure \ref{fig:ablations_checkerboard} shows our initial ablation study of four models. Three are GenAdS variants (with $\Delta = 1.5$):
\begin{itemize}
\item \textbf{AdS + KG (H)}: a residual model trained with the loss \eqref{residLoss} and the Hermite path \eqref{hermitePath};
\item \textbf{AdS + KG (L)}: a residual model trained with the loss \eqref{residLoss} and the linear path \eqref{linearPath}; and,
\item \textbf{AdS}: a model trained with the full loss \eqref{fullLoss} and the linear path \eqref{linearPath}.
\end{itemize}
The fourth, the \textbf{Baseline FCN}, is a fully-connected network with 10,601,588 parameters that performs flow matching without any AdS information. This network uses a standard linear path and an unweighted $L^{2}$ loss, without the warped-volume weighting of \eqref{fullLoss} or the Klein--Gordon backbone of \eqref{residLoss}. For a fair comparison the raw data is still mapped to the field $\tilde{\Phi}$, with $\tilde{\Pi}$ treated as ancillary.

These models probe three design axes that we treat as independent. \emph{Axis 1 (geometric):} the holographic encoding of Section \ref{subsec:holoEncod} together with the warped loss \eqref{fullLoss}, which use AdS geometry. \emph{Axis 2 (dynamical):} the Klein--Gordon backbone velocity in the residual loss \eqref{residLoss}, which uses scalar dynamics. \emph{Axis 3 (interpolation):} the linear versus cubic Hermite path, which is an ansatz-level choice within flow matching and is AdS-agnostic. The four models occupy meaningful cells of this decomposition: the \textbf{Baseline FCN} is off on all three axes; the \textbf{AdS} model is on Axis 1 only; \textbf{AdS + KG (L)} is on Axes 1 and 2; and \textbf{AdS + KG (H)} is on all three. In particular, the \textbf{AdS} model serves as the covariantised linear control, as it isolates the contribution of AdS geometry from both the scalar dynamics and the choice of path ansatz.

The four models are comparable on WED, so the discriminating metric here is BV. The GenAdS models all learn the checkerboard's boundaries faster than the FCN. In particular, each GenAdS model exhibits a sharp early decrease in BV that corresponds to the emergence of multimodality in the learned distribution. In contrast, the FCN does not show this feature. 

A tradeoff exists in per-epoch cost: the GenAdS models take roughly twice as long to train per epoch as the FCN. However, the appropriate measure of training efficiency is actually the threshold time $t_{\text{thr}}$ (the wall-clock time to reach a fixed quality target) rather than per-epoch time. On that metric, the GenAdS models are consistently faster than the FCN baseline.

We emphasise that this first ablation experiment is a preliminary comparison of the three GenAdS variants against a network that shares none of their physical or mathematical architecture. This is not necessarily a fair comparison between GenAdS and known ML approaches. In particular, the FCN does not make use of the Fourier representation of our data, and it is also ill-equipped to gauge the potential impact of the AdS propagator in the holographic encoding. These issues motivate a more refined set of experiments with more appropriately matched control models.

\subsection{Effects of Fourier representation and encoding filters}\label{subsec:controls}

The FCN comparison above leaves confounding variables that this subsection resolves. The GenAdS variants differ from the FCN not only in their AdS content but also in their representation of data (Fourier modes rather than raw coordinates) and their architecture (convolutional rather than fully connected). We therefore consider two additional experiments, run at $\sigma = 0.1$ so that the results are comparable with those presented in Figure \ref{fig:ablations_checkerboard}. The first is simply to compare our GenAdS models against a spectral CNN that removes all of the physics, while the second benchmarks models in which the spectral point encoding formula \eqref{trueSamples} is altered.

\paragraph{Spectral CNN control} The CNN used as the control in the first experiment shares everything with the network used in our GenAdS models, that is, the $16\times16$ Fourier grid, the $(\tilde{\Phi},\tilde{\Pi})$ phase space, the convolutional architecture, optimiser, and training budget. However, we strip out all three AdS ingredients, using no propagator envelope (the source modes enter unfiltered), no warped loss (plain $L^2$), and no Klein--Gordon backbone, and we also use a standard linear path. Differences between this control and the FCN from the previous experiment isolate the effect of the spectral representation and convolutional architecture, and differences among the matched linear-path GenAdS models relative to this control isolate the contribution of the AdS-specific envelope and loss structure.

\paragraph{Matched generic filters} The second experiment of this section asks whether the specific shape of the AdS envelope matters or only the use of coarse-to-fine smoothing. We replace the Bessel factor in \eqref{trueSamples} by two generic low-pass filters in the same scale variable $\xi \equiv |\vec{k}|e^{-r}$. These are a heat kernel $e^{-a\xi^2}$ and a Mat\'ern-type filter of fixed order, each with parameters fitted by least squares to the AdS envelope over the $(r,\vec{k})$ grid actually used in training (fitted scale $a = 0.398$, length $\ell = 0.636$, root-mean-square mismatch $\approx 0.05$). Everything else, including the warped loss and the shared radial schedule, is unchanged. At $\Delta = 1.5$ the comparison is particularly clean, because $\nu = 1/2$ makes the AdS envelope \eqref{kappaTilde} exactly exponential, $\tilde{\kappa} = e^{-\xi}$, so the three filters differ only in tail weight at matched scale. We also consider the case of no envelope, with just the warped loss.

\begin{table}[!ht]
\centering

\begin{tabular}{lccc}
\toprule
Model ($\sigma = 0.1$) & BV & WED & seeds\\
\midrule
AdS + KG (H) & $0.0374 \pm 0.0020$ & $0.0320 \pm 0.0086$ & 3\\
AdS + KG (L) & $0.0412 \pm 0.0025$ & $0.0247 \pm 0.0088$ & 3\\
AdS & $0.0433 \pm 0.0014$ & $0.0326 \pm 0.0045$ & 3\\
Baseline FCN & $0.0836 \pm 0.0016$ & $0.0353 \pm 0.0068$ & 3\\
Spectral CNN (no envelope, plain loss) & $\boldsymbol{0.0250 \pm 0.0017}$ & $\boldsymbol{0.0139 \pm 0.0036}$ & 3\\
Heat filter (warped loss) & $0.0412 \pm 0.0014$ & $0.0253 \pm 0.0047$ & 5\\
Mat\'ern filter (warped loss) & $0.0430 \pm 0.0010$ & $0.0292 \pm 0.0042$ & 5\\
No envelope (warped loss) & $0.0309 \pm 0.0021$ & $0.0211 \pm 0.0093$ & 5\\
\bottomrule
\end{tabular}

\caption{\justifying Matched controls at the lift-noise setting $\sigma = 0.1$. The first four rows are the models of Figure \ref{fig:ablations_checkerboard}, and the last four are the control models of this subsection. All spectral models share the same Fourier grid, phase space, architecture, optimiser, and training budget.}
\label{tab:controls}
\end{table}

\paragraph{Results} Table \ref{tab:controls} carries three conclusions, which we list now.
\begin{itemize}
\item The change in architecture from an FCN to a CNN using the Fourier representation alone accounts for the gap seen in our preliminary experiment, since the spectral CNN control, with no AdS content whatsoever, achieves the best BV and WED of any model.

\item The impact of a coarse-to-fine smoothing filter is not shape-specific, since the matched heat and Mat\'ern filters land within the same band as the AdS-enveloped linear models.

\item We can isolate the spectral control's BV advantage over \textbf{AdS} ($0.0433 \to 0.0250$) into two pieces. Removing just the envelope at fixed warped loss ($0.0433 \to 0.0309$) accounts for approximately two-thirds of the improvement, and removing the loss weighting from the model with no envelope ($0.0309 \to 0.0250$) contributes the remaining third. On this sharp-edged, high-wavenumber target, coarse-to-fine smoothing of any shape costs boundary sharpness for linear-path models, and the volume weighting of the loss carries a smaller cost of its own.
\end{itemize}
To summarise, while a natural first reading of our preliminary ablation assigns the large gain of the GenAdS models over the FCN to Axis 1 (the geometric design axis corresponding to the use of spectral encoding and a warped loss), the more refined experiments here show that this gain is carried by the change of representation and architecture through which Axis 1 was implemented, not by its AdS-specific content. However, all of the experiments so far were performed with noise injected in the momentum channel, so now we ablate that noise term.

\subsection{Lift-noise ablation}\label{subsec:liftnoise}

\begin{figure}[!ht]
\centering

\includegraphics[width=0.7\linewidth]{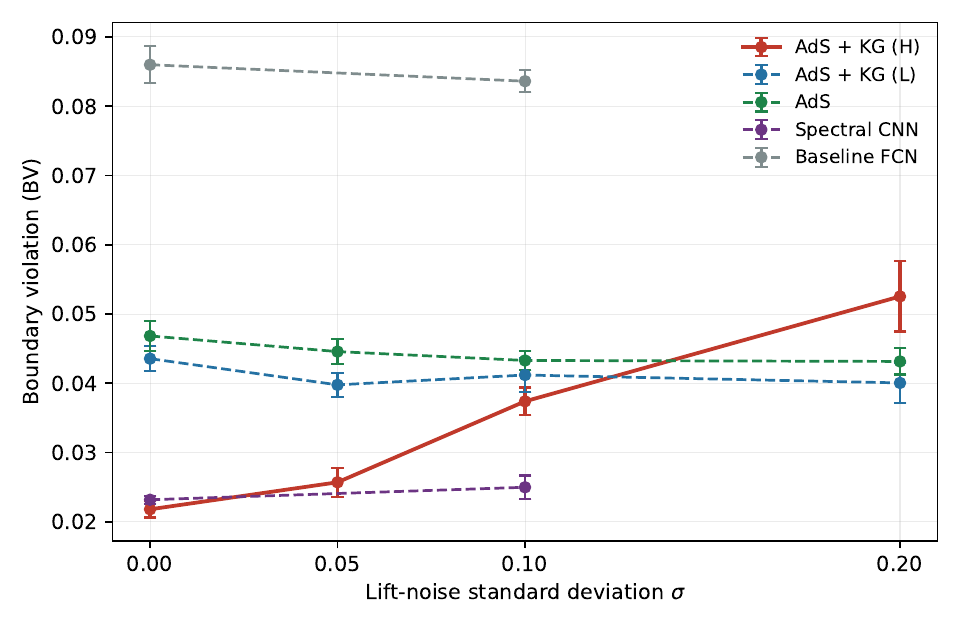}

\caption{\justifying Boundary violation as a function of the momentum-channel noise $\sigma$ for the three GenAdS variants (five seeds per point except $\sigma = 0.1$, three seeds) and the physics-free controls at the two default settings ($\sigma = 0$ and $\sigma = 0.1$). Only the Hermite-path model responds to the noise. Every model whose training targets do not consume $\tilde{\Pi}$ is flat within uncertainty.}
\label{fig:liftNoise}
\end{figure}

The experiments above at $\sigma = 0.1$ all include the momentum-channel noise described in Section \ref{subsec:holoEncod}. To quantify its effect, we ablate it over $\sigma \in \{0, 0.05, 0.1, 0.2\}$ for the three GenAdS variants, with the physics-free controls at $\sigma = 0$ and $\sigma = 0.1$, and measure BV. Table \ref{tab:lift_noise_grid} and Figure \ref{fig:liftNoise} give the result.

\begin{table}[!ht]
\centering

\begin{tabular}{lcccc}
\toprule
BV & $\sigma = 0$ & $\sigma = 0.05$ & $\sigma = 0.1^{\dagger}$ & $\sigma = 0.2$\\
\midrule
AdS + KG (H) & $\boldsymbol{0.0218 \pm 0.0012}$ & $0.0257 \pm 0.0021$ & $0.0374 \pm 0.0020$ & $0.0525 \pm 0.0051$\\
AdS + KG (L) & $0.0436 \pm 0.0018$ & $0.0398 \pm 0.0017$ & $0.0412 \pm 0.0025$ & $0.0401 \pm 0.0029$\\
AdS & $0.0468 \pm 0.0022$ & $0.0446 \pm 0.0018$ & $0.0433 \pm 0.0014$ & $0.0432 \pm 0.0019$\\
Baseline FCN & $0.0860 \pm 0.0027$ & & $0.0836 \pm 0.0016$ & \\
Spectral CNN & $0.0232 \pm 0.0006$ & & $0.0250 \pm 0.0017$ & \\
\bottomrule
\end{tabular}

\caption{\justifying Boundary violation across the lift-noise grid. Cells at $\sigma = 0.1$ (marked $\dagger$) use three seeds; all other cells use five. Like the baseline models, the alternate filter models, which all use five seeds, are consistent across $\sigma$ (heat filter, $0.0428 \pm 0.0022$ at $\sigma = 0$ versus $0.0412 \pm 0.0014$ at $\sigma = 0.1$; Mat\'ern filter, $0.0437 \pm 0.0016$ versus $0.0430 \pm 0.0010$).}
\label{tab:lift_noise_grid}
\end{table}

Across the fifteen independent Hermite-path runs at $\sigma \in \{0, 0.05, 0.2\}$, BV rises monotonically with $\sigma$ (Spearman $r = 0.93$, $p = 7\times10^{-7}$), increasing overall by a factor of $2.4$. Meanwhile, BV in all other model variants remains flat within uncertainty as we dial $\sigma$.

That the noise only affects the Hermite model can be understood from the theory in Section \ref{sec:design}. For every linear-path model, the noise perturbs a quantity that enters the regression targets only as additive zero-mean error, which the conditional-mean fit averages out. For the Hermite path, $\tilde{\Pi}_1$ sets the endpoint slope $m_1$ of the interpolant itself, so the noise deforms the trajectory that the network is trained to follow. Put another way, the noise corrupts the targets, not merely the inputs. The noise term was therefore selectively suppressing the variant that uses the most physics.

Given this result, we now restate the full comparison across all model variants at $\sigma = 0$. We write the various metrics for these models in Table \ref{tab:sigma0_checkerboard}, including coverage, which is defined as the fraction of real samples whose neighbourhoods contain a generated point. Three statements survive statistical scrutiny, and we phrase them with the corresponding tests.
\begin{itemize}
\item The Hermite-path GenAdS model decisively outperforms its linear-path counterparts, with a BV of $0.0218 \pm 0.0012$ that is roughly half of those of the linear GenAdS variants $0.0436$ to $0.0468$ (Welch $p < 10^{-6}$). So at $\sigma = 0$, the volume-adapted path fully compensates the smoothing cost that the envelope imposes on linear-path models.

\item The comparison of the Hermite-path model against the strongest physics-free control, the spectral CNN, leads to a split decision rather than a win. On BV, no statistically significant difference was detected between the two, with the Hermite model having a nominal advantage ($0.0218$ versus $0.0232$, two-sided Welch $p = 0.08$). However, the CNN retains a significant advantage on within-cell placement (WED $0.0183$ versus $0.0264$, $p = 0.04$) and on coverage ($0.9836$ versus $0.9711$, $p < 10^{-3}$).

\item The effect of envelope shape remains less significant at $\sigma = 0$. Specifically, the heat filter is marginally better than the AdS envelope in the matched no-backbone configuration ($0.0428$ versus $0.0468$, $p = 0.03$, uncorrected), the Mat\'ern filter is not significantly different from either ($p = 0.06$ against AdS, $p = 0.53$ against heat), and all three sit far above the unsmoothed control.
\end{itemize}
On the checkerboard, therefore, the Fourier representation and architecture carry the large gains, coarse-to-fine smoothing of any shape costs boundary sharpness for linear paths, and the on-shell Hermite path, the one component that uses the Klein--Gordon structure beyond generic filtering, recovers that cost in full, ranking alongside the best control on the metric its construction targets.

\begin{table}[b]
\centering

\begin{tabular}{lccc}
\toprule
Model ($\sigma = 0$, 5 seeds) & BV & WED & Coverage\\
\midrule
AdS + KG (H) & $\boldsymbol{0.0218 \pm 0.0012}$ & $0.0264 \pm 0.0046$ & $0.9711 \pm 0.0021$\\
Spectral CNN & $0.0232 \pm 0.0006$ & $\boldsymbol{0.0183 \pm 0.0046}$ & $0.9836 \pm 0.0017$\\
Heat filter & $0.0428 \pm 0.0022$ & $0.0267 \pm 0.0054$ & $\boldsymbol{0.9839 \pm 0.0020}$\\
Mat\'ern filter & $0.0437 \pm 0.0016$ & $0.0289 \pm 0.0089$ & $0.9822 \pm 0.0020$\\
AdS + KG (L) & $0.0436 \pm 0.0018$ & $0.0306 \pm 0.0030$ & $0.9799 \pm 0.0020$\\
AdS & $0.0468 \pm 0.0022$ & $0.0253 \pm 0.0060$ & $0.9810 \pm 0.0017$\\
Baseline FCN & $0.0860 \pm 0.0027$ & $0.0380 \pm 0.0028$ & $0.9703 \pm 0.0036$\\
\bottomrule
\end{tabular}

\caption{\justifying The checkerboard comparison at the default $\sigma = 0$, five seeds per cell. Coverage is the fraction of real samples with a generated sample in their feature-space neighbourhood. Bold marks the best mean per column; significance is discussed in the text.}
\label{tab:sigma0_checkerboard}
\end{table}

\subsection{Physical experiments}\label{subsec:physExp}
As GenAdS is a physics-informed framework, we are motivated to also test the effect of varying physical parameters relevant to our models. We do that now so as to probe the effects of modulating two properties, the mass of the scalar and the flatness of AdS.

\paragraph{Modulating scalar masses in AdS}

\begin{figure}[!ht]
\centering
\includegraphics{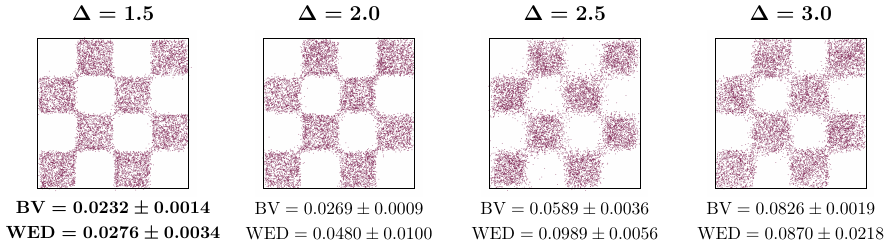}
\caption{\justifying
The results from varying the scaling dimension $\Delta$, which corresponds to the bulk scalar's squared-mass $m^2 = \Delta(\Delta-2)$. These models are trained to 100 epochs. All numbers are mean $\pm$ standard deviation over three seeds at the default $\sigma = 0$. Across the four tested values, $\Delta = 1.5$ is best on both metrics. BV also degrades monotonically with $\Delta$.}
\label{fig:scalarMasses}
\end{figure}

The scaling dimension $\Delta$ controls the bulk scalar mass through $m^2 = \Delta(\Delta - d)$, which for $d = 2$ reads $m^2 = \Delta(\Delta-2)$. It is natural to ask how $\Delta$ affects the model. We probe this by plugging in the following values:
\begin{equation}
\Delta = 1.5,2,2.5,3 \implies m^2 = -0.75,0,1.25,3.
\end{equation}
The results are shown in Figure \ref{fig:scalarMasses}. At the default $\sigma = 0$ and with three seeds per value, $\Delta = 1.5$ is optimal on both metrics, and BV degrades monotonically across the four tested values, with every pairwise difference significant (Welch $p \leq 0.045$). We restrict this statement to the tested range and make no claim about a continuous mass dependence.

We stress that the ranking of $\Delta$ values depends on the lift-noise setting; at $\sigma = 0.1$ the ordering differs, with $\Delta = 2.0$ best on BV and $\Delta = 1.5$ best on WED (BV $0.0349 \pm 0.0004$ at $\Delta = 2.0$; WED $0.0370 \pm 0.0041$ at $\Delta = 1.5$; three seeds each). This $\sigma$-dependence is itself informative, since the noise couples to the model through the $\Delta$-dependent momentum envelope.

This $\Delta$-dependence can be understood by analysing the stabilised Klein--Gordon equations \eqref{kg1stabl}--\eqref{kg2stabl}. In the near-boundary regime where $|\vec{k}|^2 e^{-2r} \ll 1$, the second of these equations reduces to $d\tilde{\pi}_{\vec{k}}/dr = -(2\Delta-d)\tilde{\pi}_{\vec{k}}$. So, the radial dynamics has characteristic exponent $-(2\Delta - d)$, and the KG backbone $V_{\text{KG}}$ thus carries a damping rate of $2\Delta-d$. The residual loss \eqref{residLoss} then requires the neural residual $R_t(\theta)$ to cancel this backbone down to the flow-matching target velocity, a cancellation whose scale grows linearly with $(2\Delta-d)$ and is correspondingly harder to learn as $\Delta$ increases. That said, we merely offer this argument as a heuristic for the degradation trend, not a derivation, since the trend involves the full $\Delta$-dependent stack.

\paragraph{Interpolating between flat space and AdS}

Our machinery can be adapted to any geometry, which raises the question of where AdS sits within the larger design space. We probe this by scanning through the hyperscaling-violating (HSV) family of geometries \cite{Huijse:2011ef}. These are a one-parameter class of metrics that interpolate between AdS and flat space, with AdS at the conformal endpoint at which the boundary theory has no anomalous scaling. Our scan moves off of this endpoint in a controlled manner. The HSV metric is
\begin{equation}
ds^2 = dr^2 + \frac{1}{(pr)^{2\gamma}} d\vec{x}^2,\ \ \gamma \equiv \frac{1}{p}-1,\label{hsvMetDW}
\end{equation}
where $p \in (0,1]$ is the interpolation parameter and the boundary sits at $r = 0$. Setting $p \to 1$ ($\gamma = 0$) recovers flat space, while $p \to 0$ ($\gamma \to \infty$) is the AdS limit.

We again represent the data with scalar fields, but HSV imposes a restriction not present in AdS: the Klein--Gordon equation \eqref{kleinGordonWarped} admits a closed-form solution only for $m^2 = 0$. The relevant formulas are collected in Appendix \ref{app:HSVdisc}.

\begin{figure}[!ht]
\centering
\includegraphics{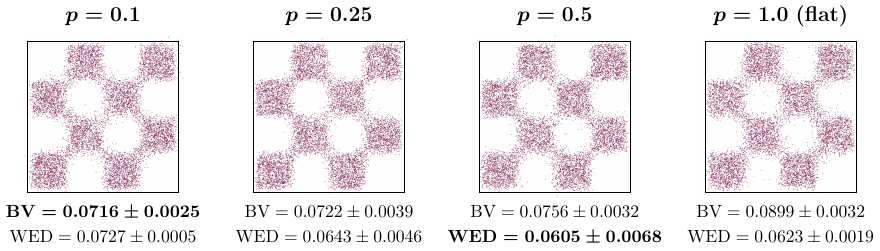}
\caption{\justifying
The results from dialing the HSV parameter $p$. These models have $m^2 = 0$ and are trained to 100 epochs. All numbers are mean $\pm$ standard deviation over three seeds at the default $\sigma = 0$.}
\label{fig:hsvFits}
\end{figure}

We train massless HSV models with $p \in \{0.1,\,0.25,\,0.5,\,1.0\}$ on the checkerboard. To make the comparison fair, the radial cutoffs are chosen so that the coordinate $z \equiv (pr)^{1/p}$ entering the propagator \eqref{propagatorHSV} satisfies $z_{\text{IR}} = 1$ and $z_{\text{UV}} = 1/e$, matching the GenAdS choice ($z = e^{-r}$). The results are shown in Figure \ref{fig:hsvFits}.

With three seeds per value at the default $\sigma = 0$, the comparison is unambiguous. The planar AdS model outperforms every tested HSV configuration simultaneously on both metrics. The best HSV cell reaches BV $0.0716 \pm 0.0025$ and WED (at a different $p$) $0.0605 \pm 0.0068$, against $0.0218 \pm 0.0012$ and $0.0264 \pm 0.0046$ for the planar Hermite model of Table \ref{tab:sigma0_checkerboard}, a factor of three on BV with Welch $p = 3\times10^{-4}$ against the best HSV member. Within the family, BV worsens from the AdS end towards flat space across the tested values (Spearman $p = 0.005$ over the twelve runs), while WED is non-monotonic with its best values at mid-to-large $p$.

Two caveats apply. There is room for hyperparameter tuning, particularly with the cutoff values. Additionally, the AdS endpoint $p \to 0$ is a singular limit of this parameterisation (the $r$-coordinate blows up there, as noted in Appendix \ref{app:HSVdisc}), so no finite-$p$ member of the scan is expected to coincide with the planar AdS pipeline. The scan probes departure from AdS, not convergence to it.

\section{Generating MNIST}\label{sec:experimentMNIST}

The Modified National Institute of Standards and Technology (MNIST) dataset of hand-drawn digits \cite{lecun2002} is a standard benchmark for image generation. It is simple by image-dataset standards, but its dimensionality dwarfs that of the checkerboard: each image carries $28^2 = 784$ pixels. MNIST therefore tests whether GenAdS scales to higher-dimensional data.

We continue to work in AdS$_3$ ($d = 2$) with $\Delta = 1.5$. The encoding, however, changes; instead of $\delta$ sources, we use the pixel intensity maps as boundary sources. The velocity CNN has 13,448,514 parameters for the Hermite path and 13,449,668 for the linear path, somewhat larger than in the checkerboard experiments. Its architecture follows the same 2D convolutional template as in Section \ref{sec:experimentsCheckerboard}, scaled up in width; the full specification is documented in the accompanying code repository. Training uses 10,000 samples per epoch with batch size 128, AdamW with learning rate $3 \times 10^{-4}$ and weight decay $1 \times 10^{-5}$. The base distribution parameters are those of \eqref{paramsIRDist}. The spectral parameters are $L = K = 28$, giving 784 Fourier modes. The lift-noise parameter is set to both $\sigma = 0.1$ and $\sigma = 0$. All reported numbers are means $\pm$ standard deviations over three seeds, and each run records its configuration, command line, and commit.

\subsection{MNIST metrics}

Fr\'echet inception distance (FID) \cite{heusel2018ganstrainedtimescaleupdate} is the standard metric for image-generation quality on datasets like CIFAR-10. FID compares the ground-truth and model distributions in the feature space of a pretrained Inception network \cite{szegedy2015rethinkinginceptionarchitecturecomputer}, capturing semantic differences rather than pixel-level ones.

Concretely, the feature-space distributions are modelled as multivariate Gaussians. For a set of $N$ images $\{X_i\}$ and learned features $\vec{\mathcal{F}}(X_i)$, the mean and covariance are
\begin{align}
\vec{\mu} &\equiv \frac{1}{N}\sum_{i=1}^N \mathcal{F}(X_i),\\
\Sigma &\equiv \frac{1}{N-1}\sum_{i=1}^N \left[\vec{\mathcal{F}}(X_i) - \vec{\mu}\right] \left[\vec{\mathcal{F}}(X_i) - \vec{\mu}\right]^T.
\end{align}
The FID is the 2-Wasserstein distance between the resulting Gaussians $\mathcal{N}(\vec{\mu}_r,\Sigma_r)$ for real and $\mathcal{N}(\vec{\mu}_g,\Sigma_g)$ for generated data, which has the closed form
\begin{equation}
\text{FID} = \left\|\vec{\mu}_{r} - \vec{\mu}_{g}\right\|^2 + \text{Tr}\left[\Sigma_{r} + \Sigma_{g} - 2\left(\Sigma_{r} \Sigma_{g}\right)^{1/2}\right].
\end{equation}
A vanishing FID means the two distributions agree; lower FID means better agreement.

In our examination, we report three further metrics beyond Inception-space FID, so that fidelity and coverage can be assessed separately. The first is the kernel inception distance (KID) \cite{Binkowski:2018kid}, an unbiased squared maximum-mean-discrepancy estimate with a cubic polynomial kernel, which avoids the Gaussian assumption and finite-sample bias of FID. The second and third, respectively, are improved precision and recall \cite{Kynkaanniemi:2019pr}, which use $k$-nearest-neighbour-manifold estimates to separate sample fidelity from coverage; we use $k=3$ and construct these manifolds in Inception space.

The evaluation protocol is fixed across models, with $10{,}000$ generated samples per seed at deterministic seeds compared against a seeded $10{,}000$-image reference drawn from the training split. Recomputing every metric against the held-out test split changes the values by at most a few tenths and no ordering, so the conclusions below are robust to the choice of reference.

\subsection{Ablations with MNIST}

We repeat the ablation study of Section \ref{subsec:ablation_checkerboard} on MNIST. The three GenAdS variants are unchanged: \textbf{AdS + KG (H)} (residual loss \eqref{residLoss} with Hermite path \eqref{hermitePath}), \textbf{AdS + KG (L)} (residual loss with linear path \eqref{linearPath}), and \textbf{AdS} (full loss \eqref{fullLoss} with linear path). This time, the only baseline is a CNN, since CNNs are far better suited to image generation through their inductive bias towards translational equivariance. The baseline shares the architecture of the linear-path GenAdS CNNs, with 13,449,668 parameters. Training times are uniform across models in this experiment, so we report performance by epoch rather than by wall-clock time.

Table \ref{tab:mnist_extended} reports the $\sigma = 0.1$ results with just FID and the $\sigma = 0$ results with the extended metric suite. Starting with the former, the $\sigma = 0.1$ results read cleanly through the three-axis decomposition of Section \ref{subsec:ablation_checkerboard}. The \textbf{AdS} model ($22.70 \pm 1.66$; Axis 1 only) and \textbf{AdS + KG (L)} ($21.76 \pm 2.02$; Axes 1 and 2) are both within one standard deviation of the \textbf{Baseline CNN} ($21.99 \pm 2.90$). So on MNIST, the AdS-specific axes incur no detectable FID cost. In contrast, \textbf{AdS + KG (H)} ($38.17 \pm 0.45$; all three axes) is several standard deviations worse.

\begin{table}[!ht]
\centering

\begin{tabular}{lcccc}
\toprule
Model ($\sigma = 0.1$) & FID & & &\\
\midrule
AdS + KG (H) & $38.170 \pm 0.449$ & & &\\
AdS + KG (L)  & $\boldsymbol{21.762 \pm 2.021}$ & & &\\
AdS & $22.703 \pm 1.659$ & & &\\
Baseline CNN  & $21.991 \pm 2.897$ & & &\\
\bottomrule
\toprule
Model ($\sigma = 0$) & FID & KID $\times 10^3$ & Precision & Recall\\
\midrule
AdS + KG (H) & $27.86 \pm 0.82$ & $19.32 \pm 2.05$ & $0.964 \pm 0.019$ & $0.319 \pm 0.012$\\
AdS + KG (L) & $16.93 \pm 0.50$ & $5.29 \pm 0.46$ & $0.974 \pm 0.007$ & $0.891 \pm 0.017$\\
AdS & $17.33 \pm 0.35$ & $6.66 \pm 0.47$ & $0.958 \pm 0.019$ & $\boldsymbol{0.896 \pm 0.011}$\\
Baseline CNN & $\boldsymbol{16.20 \pm 0.68}$ & $\boldsymbol{3.81 \pm 1.18}$ & $\boldsymbol{0.984 \pm 0.013}$ & $0.824 \pm 0.016$\\
\bottomrule
\end{tabular}

\caption{\justifying The MNIST comparison at the default $\sigma = 0$, three seeds per model, reporting Inception-space FID, KID, and improved precision and recall, computed from $10{,}000$ generated samples per seed against a seeded $10{,}000$-image training-split reference. We also report FID for a model comparison with lift-noise activated ($\sigma = 0.1$). Bold marks the best mean per column.}
\label{tab:mnist_extended}
\end{table}

\begin{figure}[!ht]
\centering
\includegraphics{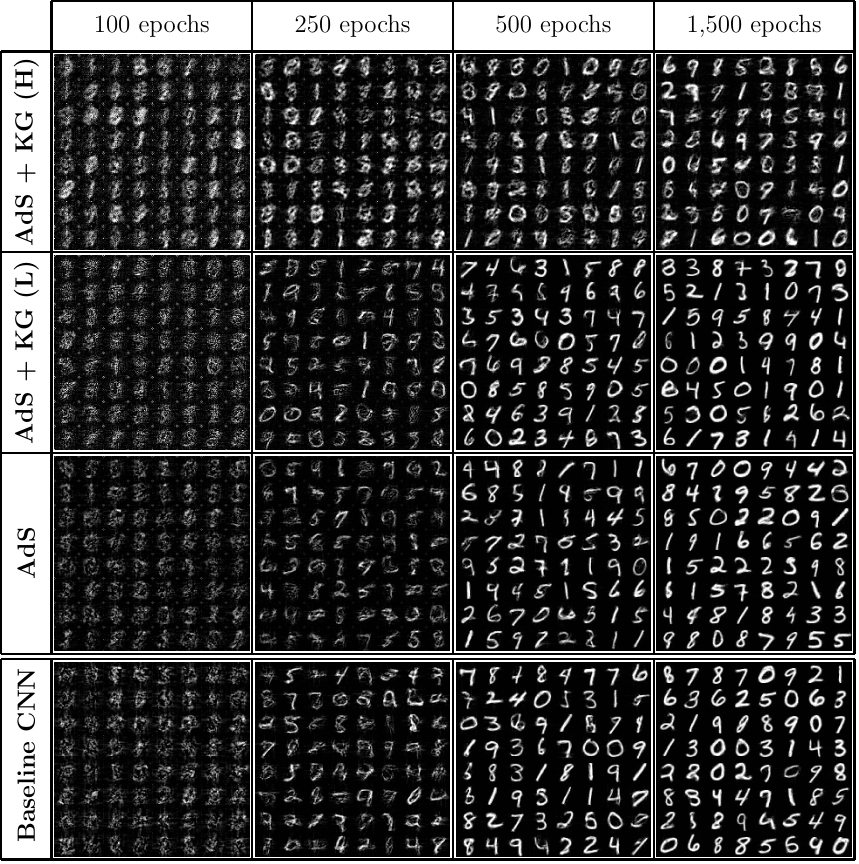}
\caption{\justifying
The ablation experiment for MNIST. We present samples generated by four different models, \textbf{AdS + KG (H)}, \textbf{AdS + KG (L)}, \textbf{AdS}, and \textbf{Baseline CNN}, after 100, 250, 500, and 1500 training epochs. The first three models use AdS physics, with \textbf{AdS + KG (H)} trained on the residual loss \eqref{residLoss} with a Hermite path \eqref{hermitePath}, \textbf{AdS + KG (L)} trained on the residual loss with a linear path \eqref{linearPath}, and \textbf{AdS} trained to learn the full velocity with the loss \eqref{fullLoss} with a linear path. The \textbf{Baseline CNN} is physics-free but is still trained using flow matching with a linear path. The samples shown are from the $\sigma = 0$ runs, whose metrics are given in Table \ref{tab:mnist_extended}.}
\label{fig:ablations_mnist}
\end{figure}

The $\sigma = 0$ results both confirm and refine this reading; generated samples after 100, 250, 500, and 1500 epochs are shown in Figure \ref{fig:ablations_mnist}. Removing the lift noise improves every linear model by roughly five FID points and the Hermite model by ten ($38.17 \to 27.86$). That the largest gain is seen in the Hermite model is consistent with the results of Section \ref{subsec:liftnoise}, through which we concluded that the noise most strongly corrupts the Hermite path, though it still loses to the linear models on all metrics. However, the ordering among the linear-path models is unchanged; the CNN baseline holds a small nominal edge on FID and KID, with \textbf{AdS + KG (L)} and \textbf{AdS} close behind. The precision-recall decomposition, however, reveals a genuine trade-off that FID alone conceals; the AdS-encoded linear models achieve significantly higher recall than the baseline ($0.891 \pm 0.017$ and $0.896 \pm 0.011$ versus $0.824 \pm 0.016$, roughly four standard errors) at comparable precision, so the holographic encoding provides measurably better coverage of the digit manifold at a small cost in Fr\'echet statistics.

The extended metrics also convert the Hermite outlier from a number into a diagnosis. Its precision is high ($0.964 \pm 0.019$) while its recall collapses to $0.319 \pm 0.012$. So, the model produces class-consistent digits that occupy only a narrow region of the data manifold. The Inception-space FID and KID, together with the pixel statistics (generated standard deviation $0.53$ against $0.62$ for the data), register an accompanying loss of low-level fidelity, and the sample grids of Figure \ref{fig:ablations_mnist} show both effects directly. A classifier-feature FID, which uses the same Fr\'echet construction as Inception-space FID but is instead evaluated in the feature space of a small convolutional digit classifier trained on MNIST (test accuracy $0.988$), corroborates the Inception ordering; the classifier-feature FID values are $138$, $174$, $187$, $428$ for the baseline, linear, AdS, and Hermite models respectively. The Hermite path therefore induces a characterised coverage collapse with degraded low-level statistics, which is precisely the feature that decisively helps on the sharp-edged, low-dimensional checkerboard. Path choice, in short, is task-dependent.

\vfill
\section{Conclusions and outlook}\label{sec:conclusion}

GenAdS provides a concrete implementation of holography-inspired generative modelling. The construction we have developed is deliberately simple, but already the experiments reveal several features worth recording.

The experiments, controls, and ablations together support the following picture. On the checkerboard, the large advantage of the GenAdS variants over a raw-coordinate FCN is carried by the Fourier representation and convolutional architecture, as the matched spectral control of Section \ref{subsec:controls} demonstrates. The AdS envelope itself acts as generic smoothing, replicable by fitted heat and Mat\'ern filters, and it costs boundary sharpness for linear-path models, as does the warped loss to a lesser degree.

The clearest benefit associated with the Klein--Gordon structure is obtained when it is used to construct the on-shell Hermite path. The lift-noise ablation of Section \ref{subsec:liftnoise} shows that the momentum regularisation selectively suppresses exactly this variant. As the noise is dialled, the Hermite model exhibits the only clear monotonic change in a seven-model grid. With noise set to zero, the Hermite model halves boundary violation relative to the linear-path variants; on that metric, no statistically significant difference was detected between the Hermite model and the strongest physics-free control, the spectral CNN (two-sided Welch $p = 0.08$).

On MNIST, GenAdS delivers a measured trade-off, with the linear models achieving significantly higher recall than the CNN baseline in exchange for a small nominal cost in Fr\'echet statistics. However, the Hermite model fails due to a coverage collapse.

The implementation of holographic encoding is itself a place where refinement is possible, particularly for images. One natural variant would represent each pixel as a small collection of spatially arranged point sources, generalising the point encoding to higher-dimensional data. This would also call for a more refined decoding procedure than the one we use here.

The GenAdS model is built on AdS physics from the outset, so the experimental results admit a natural physics interpretation. We present this as a characterisation of the physics-informed model, not as an independent physics result. The $\Delta$ scan covers three qualitatively different cases, namely $1 < \Delta < 2$, $\Delta = 2$, and $\Delta > 2$, which for $d = 2$ correspond to negative, zero, and positive $m^2$. On the boundary side, these correspond to relevant, marginal, and irrelevant deformations \cite{Aharony:1999ti}. The intuition is that $\Delta > 2$ scalars are irrelevant in the IR (deep in the bulk), and the model is correspondingly less stable. At $\sigma = 0$, the $\Delta$ scan gives this intuition support within the tested range. $\Delta = 1.5$, the relevant-operator case, is best on both metrics; BV worsens monotonically over the four tested values, while WED does not.

The hyperscaling-violating experiments scan one axis of departure from AdS, the degree of anomalous scaling of the boundary theory. Within the tested configurations, the planar AdS model outperforms every HSV member simultaneously on both metrics, including by a factor of three on BV at high significance (Section \ref{subsec:physExp}). The HSV framework is also more restrictive than AdS because the Klein--Gordon equation admits closed-form solutions only for massless scalars in HSV. Accordingly, the preference for $m^2<0$ indicated by the $\Delta$ scan is restricted to the AdS configurations tested here.

Several directions for future work suggest themselves. The first is non-Euclidean data. The present treatment uses planar AdS only; AdS in spherical or hyperbolic slicings is equally available, and a spherical version of GenAdS would be a natural arena for generating data on a sphere.

A second direction is to look beyond physical equations of motion to the geometric structure of AdS itself. AdS geometry has long been known, in the AdS/CFT community, to encode universal information about the boundary; the most familiar instance is the Ryu--Takayanagi relationship between CFT entanglement entropy and minimal-area surfaces in the bulk \cite{Ryu:2006bv}. More broadly, AdS/CFT provides a growing family of computable geometric observables that encode boundary-theoretic information content, and incorporating such observables as inductive biases for generation is a natural direction for future work on GenAdS. A programme running in the \emph{inverse} direction uses deep learning to infer holographic bulk structure from boundary data \cite{Hashimoto:2018ftp,PhysRevD.99.106017}; GenAdS is the \emph{forward} counterpart, using holographic structure as an inductive bias for machine learning rather than using machine learning to learn holography.

A related limitation is that the present bulk theory is free. Under the AdS/CFT dictionary, a free bulk scalar is dual to a generalised free boundary field whose connected $n$-point functions vanish for $n \neq 2$, so all higher-point correlators factorise into Wick contractions of the two-point function \cite{Aharony:1999ti}. The holographic encoding of Section \ref{subsec:holoEncod} therefore carries only two-point content, so any non-Gaussian structure in the generated samples is supplied by the neural-network flow rather than by the encoding itself. The most direct holographic extension is to promote the bulk to an interacting theory. Bulk interaction vertices generate tree-level Witten-diagram contributions to connected boundary $n$-point functions for $n \geq 3$ \cite{Witten:1998qj,Aharony:1999ti}, giving the encoding non-Gaussian content and modifying the target velocity in the residual loss \eqref{residLoss} accordingly.

That said, we might question whether Witten-diagram vertices supply genuinely useful non-Gaussian inductive bias (structure the residual network would otherwise have to learn, delivered analytically at known coupling) or whether, as the free-field controls here might suggest, a sufficiently expressive residual renders the analytic non-Gaussianity redundant. A $\lambda\Phi^3$ or $\lambda\Phi^4$ bulk vertex gives the leading tractable test, since the encoding then acquires a computable connected three-point or four-point kernel. The free-field ceiling of this paper is thus not merely a limitation but a measured baseline against which interacting extensions can be judged.

Finally, a more ambitious goal is to incorporate gravitational backreaction. Such a model would account for the response of the bulk metric to the scalar dynamics. The flow equations involved are considerably more complicated than Klein--Gordon, since they include first and second derivatives of the metric, but they also have natural interpretations in the language of renormalisation-group flow, which has been linked to optimal transport and diffusion \cite{Cotler:2022fze,Cotler:2023lem}. GenAdS with backreaction could thus serve as a geometric realisation of the connection between generative modelling and RG flow.

GenAdS is a new implementation of holography-inspired generative modelling. It provides a working construction whose inductive bias is analytic and auditable, and it is ripe for extension in a myriad of directions. Our experiments show that ideas from quantum gravity and holography, abstract though they may seem, can be built into machine-learning architectures and characterised with the same rigour as any other design choice.

\funding{SS and ES were supported by the U.S. Department of Energy (DOE) grant DE-SC001010. EM, SS, and RM were supported by the National Science Foundation STTR award 2451680. DV was supported by DOE grant DE-SC0024563. RM was also supported by NSF Maximize ACCESS allocation number BIO220163 and the DOE National Energy Research Scientific Computing Center (NERSC).}

\data{The code is publicly available at our \href{https://github.com/LVCIFERI/Holographic-Generative-Flows-with-AdS-CFT---Planar}{GitHub repository}. The results of this paper correspond to the tagged release \texttt{v1.1-mlst-revision}, commit \texttt{45b7326}, archived at \href{https://doi.org/10.5281/zenodo.21887138}{doi:10.5281/zenodo.21887138}.
 Every reported experiment ships with its exact configuration file, command line, and git commit, recorded automatically alongside the run outputs, together with the per-seed metrics and aggregation scripts. The MNIST dataset is publicly available. The checkerboard distribution is generated synthetically as described in the text.}
\vfill\pagebreak
\begin{appendix}
\section{Hyperscaling-violating geometries}\label{app:HSVdisc}

The HSV experiments of Section \ref{subsec:physExp} require some additional setup, namely rewriting the metric in warped form, deriving the propagator, and recording the asymptotic behaviour of the scalar. We collect these formulas here.

The metric is most naturally written in terms of a coordinate $z$,
\begin{equation}
ds^2 = \frac{1}{z^{2(1-p)}}\left(dz^2 + d\vec{x}^2\right),\ \ z > 0,\ \ p \in (0,1].\label{hsvmetric1}
\end{equation}
The slice $z = 0$ is special. In the AdS limit $p \to 0$ it is the location of the dual CFT, while in the flat limit $p \to 1$ it is an artificial coordinate boundary. For interpolating values of $p$ it is the slice on which the metric diverges.

\subsection{Rewriting the metric}

For $p \in (0,1]$, the substitution $r \equiv z^p/p$ is well defined on $z > 0$ and brings \eqref{hsvmetric1} into the warped form \eqref{warpedAnsatz},
\begin{equation}
ds^2 = dr^2 + \frac{1}{(pr)^{2\gamma}} d\vec{x}^2,\ \ \gamma \equiv \frac{1}{p}-1 \in [0,\infty).
\end{equation}
The transverse slices are scaled by a power of $r$ rather than an exponential, in contrast to AdS. The AdS limit $p \to 0$ is also where this $r$-coordinate develops a coordinate singularity. Note further that $r$ now runs on $(0,\infty)$ with $r = 0$ as the boundary, opposite to the AdS convention; the cutoffs accordingly satisfy $r_{\text{UV}} < r_{\text{IR}}$.

The path is again written in the cubic Hermite basis with $u \in [0,1]$ defined by \eqref{affineParams}, but the explicit form of $u(r)$ is now
\begin{equation}
u(r) = \frac{r_{\text{IR}}^{d\gamma + 1} - r^{d\gamma + 1}}{r_{\text{IR}}^{d\gamma + 1} - r_{\text{UV}}^{d\gamma + 1}}.\label{pathHSV}
\end{equation}

\subsection{Solving Klein--Gordon}

We turn to the propagator. The scalar field admits the Fourier decomposition
\begin{equation}
\Phi(r,\vec{x}) = \int \frac{d^d k}{(2\pi)^{d/2}} e^{-i\vec{k} \cdot \vec{x}} \phi_{\vec{k}}(r),
\end{equation}
and the Fourier-space mode equation follows from \eqref{ODEmodesGen} on identifying $\lambda = |\vec{k}|^2$ and $f(r) = (pr)^{-\gamma}$,
\begin{equation}
\frac{d^2 \phi_{\vec{k}}}{dr^2} - \frac{d\gamma}{r} \frac{d\phi_{\vec{k}}}{dr} = \left[(pr)^{2\gamma}|\vec{k}|^2 + m^2\right]\phi_{\vec{k}}(r).\label{ODEmodes}
\end{equation}
The same equation governs the propagator coefficients in momentum space. Unlike in AdS, the massive case admits no closed-form solution \cite{Huijse:2011ef}. The massless case, however, does. Imposing regularity and the normalisation $\kappa \to 1$ as $r \to 0$, one finds
\begin{equation}
\kappa_{|\vec{k}|}(r) = \frac{\left[(pr)^{1/p}|\vec{k}|\right]^\beta}{2^{\beta-1}\Gamma(\beta)} K_\beta\left((pr)^{1/p}|\vec{k}|\right),\ \ \ \ \beta = \frac{1+(d-1)(1-p)}{2}.\label{propagatorHSV}
\end{equation}
Setting $m^2 = 0$ in \eqref{ODEmodes} and keeping the dominant terms as $r \to 0$ gives the asymptotic form
\begin{equation}
\phi_{\vec{k}} \sim J(\vec{k}) + r^{d\gamma + 1}\phi_1(\vec{k}),
\end{equation}
where $J$ is again the source and $\phi_1$ plays the role of the VEV in \eqref{asymptoticPhi}. The source is already $O(1)$ and the VEV piece decays as $r \to 0$, so no field redefinition is needed for numerical stability.

Flow matching therefore proceeds with the original $\Phi$ and its conjugate momentum $\Pi \equiv \partial_r \Phi$. In Fourier space the first-order massless system reads
\begin{align}
\frac{d{\phi}_{\vec{k}}}{dr} &= {\pi}_{\vec{k}}(r),\label{kg1HSV}\\
\frac{d{\pi}_{\vec{k}}}{dr} &=  (p r)^{2\gamma} |\vec{k}|^2 {\phi}_{\vec{k}}(r) + \frac{d\gamma}{r} {\pi}_{\vec{k}}(r).\label{kg2HSV}
\end{align}
With these in hand the construction is identical to its AdS counterpart; spectral point encoding, residual loss in Fourier space, all as in Section \ref{sec:design}, with propagator \eqref{propagatorHSV}, mode equations \eqref{kg1HSV}--\eqref{kg2HSV}, and Hermite-path parameter \eqref{pathHSV} adapted to the HSV geometry.
\end{appendix}

\end{document}